# Decentralized Supply Chain Formation: A Market Protocol and Competitive Equilibrium Analysis

**William E. Walsh**                                          WWALSH1@US.IBM.COM
*IBM T. J. Watson Research Center*
*19 Skyline Drive*
*Hawthorne, NY 10532 USA*

**Michael P. Wellman**                                          WELLMAN@UMICH.EDU
*University of Michigan AI Laboratory*
*1101 Beal Avenue*
*Ann Arbor, MI 48109-2110 USA*

## Abstract

Supply chain formation is the process of determining the structure and terms of exchange relationships to enable a multilevel, multiagent production activity. We present a simple model of supply chains, highlighting two characteristic features: hierarchical subtask decomposition, and resource contention. To decentralize the formation process, we introduce a market price system over the resources produced along the chain. In a competitive equilibrium for this system, agents choose locally optimal allocations with respect to prices, and outcomes are optimal overall. To determine prices, we define a market protocol based on distributed, progressive auctions, and myopic, non-strategic agent bidding policies. In the presence of resource contention, this protocol produces better solutions than the greedy protocols common in the artificial intelligence and multiagent systems literature. The protocol often converges to high-value supply chains, and when competitive equilibria exist, typically to approximate competitive equilibria. However, complementarities in agent production technologies can cause the protocol to wastefully allocate inputs to agents that do not produce their outputs. A subsequent decommitment phase recovers a significant fraction of the lost surplus.

## 1. Introduction

Electronic commerce technology can provide significant improvements in existing modes of commercial interaction, through increased speed, convenience, quality, and reduced costs. Yet some have proposed more radical visions of how business may be transformed. Exponential increases in communications bandwidth and computational ability have the potential to qualitatively decrease the friction in business interactions. With this as a premise, Malone and Laubaucher's treatise on the emerging "E-Lance Economy" (1998) puts forth the view that, in the not-too-distant future, business relationships will lose much of their current persistent character. Indeed, Malone and Laubaucher propose that large companies as we know them will cease to exist, and rather be dynamically formed by "electronically connected freelancers" (e-lancers) for the purpose of producing particular goods and services, and then dissolved when projects are completed. Others employ the evocative term "virtual corporation" (Davidow, 1992) to describe groups of agile organizations forming temporary confederations for ad hoc purposes.

Whether or not one accepts the full extent of this vision of virtual corporations, several business trends provide evidence that we are moving in this direction. Software companies are time-shifting development between the U.S. and India, and Sun Microsystems now allows freelance programmers to bid to fix customers' software problems (Borenstein & Saloner, 2001). Large, traditional manufacturing companies, exemplified by major automotive manufacturers, increasingly outsource the production of various components. Ford and General Motors (GM) have spun off parts manufacturing into separate companies (Lucking-Reily & Spulber, 2001). Start-ups and other small companies form partnerships to compete with larger, more established companies. Application service providers supplant in-house provision of standard operations, information, and technology services.

We study this phenomenon in the guise of supply chains, a common form of coordinated commercial interaction. For our purposes, a ***supply chain*** is a network of production and exchange relationships that spans multiple levels of production or task decomposition. Whenever we have a producer that buys inputs and sells outputs, we have a supply chain. Although typically used to refer to multi-business structures in manufacturing industries, any service or contracting relationship that spans multiple levels can be viewed as a supply chain.

***Supply chain formation*** is the process of determining the participants in the supply chain, who will exchange what with whom, and the terms of the exchanges. Traditionally, supply chains have been formed and maintained over long periods of time by means of extensive human interactions. But the acceleration of commercial decision making is creating a need for more advanced support. Companies ranging from auto makers to computer manufacturers are basing their business models on rapid development, build-to-order, and customized products to satisfy ever-changing consumer demand. And fluctuations in resource costs and availability mean that companies must respond rapidly to maintain production capabilities and profits. As these changes increasingly occur at speeds, scales, and complexity unmanageable by humans, the need for automated supply chain formation becomes acute.

Because the agents are autonomous in an electronic commerce setting, we must generally assume that they have specialized knowledge about their own capabilities but limited knowledge about other individuals and the large-scale structure of the problem. Because agents are self-interested, they will participate with the goal of maximizing their own benefit. Additionally, we may have cause to control the allocation of each resource individually if, for instance, global optimization is infeasible or if no one entity has global allocative authority. For such environments where information, decision making, and control are inherently decentralized, we seek to engineer the process of bottom-up supply chain formation. This problem is complicated if the structure of resource contention precludes the use of simple greedy allocation strategies.

We present a decentralized, asynchronous market protocol for supply chain formation under conditions of resource scarcity. The protocol allows agents to negotiate the formation of supply chains in a bottom-up fashion, requiring only local knowledge and communication. In the market protocol, agents' decisions are coordinated by the price system, with the price for each resource determined through an ascending auction.

The remainder of the paper describes our market protocol, and characterizes its behavior theoretically and empirically.[1] We begin in Section 2 with a formal definition of the supply chain formation problem, and an illustrating application to the automotive industry. In Section 3, we show how typical greedy top-down approaches to supply chain formation can fail in the presence of resource contention. We define a price system and analyze static properties of price equilibria in Section 4. In Section 5, we introduce a price-based market protocol for supply chain formation and analyze its convergence properties. We present the results of an empirical study of the protocol in Section 6. In Section 7, we discuss relevant results and issues in price-based analysis and auction theory, as well as some related work in supply chain formation. We conclude in Section 8 and suggest extensions and future work. Throughout, we defer proofs to Appendix A.

---

1. Further details may be found in the first author's dissertation (Walsh, 2001).

## 2. The Supply Chain Formation Problem

Agents in the supply chain are characterized in terms of their capabilities to perform tasks, and their interests in having tasks accomplished. A central feature of our model of the problem is ***hierarchical task decomposition***: in order to perform a particular task, an agent may need to achieve some subtasks, which may be delegated to other agents. These may in turn have subtasks that may be delegated, forming a supply chain through a decomposition of task achievement. Constraints on the task assignment arise from resource contention, where agents require a common resource (e.g., a task achievement, or something tangible such as a piece of equipment) to accomplish their tasks.

Tasks are performed on behalf of particular agents; if two agents need a task then it would have to be performed twice to satisfy them both. In this way, tasks are the same as any other discrete, rival resource. Hence, we make no distinction in our model, and use the term "good" to refer to any task or resource provided or needed by agents. The assumption that goods cannot be shared or reused (i.e., have limited available quantities) is necessary for much of our analysis. Goods that can be replicated at little or no marginal cost, such as software and information, provide many interesting challenges to economic analysis (Shapiro & Varian, 1999), not addressed in this work.

### 2.1 Example: Automotive Supply Chain Formation

We illustrate our model of supply chain formation with an application to a stylized, hypothetical example from the automotive industry. Traditionally, automotive supply chains span many tiers, formed and maintained over long periods of time through extensive human negotiations. Some automation is emerging, for example through Covisint[2], a company formed by GM, Ford, and DaimlerChrysler to mediate the negotiation and exchange of parts, as well as other supply chain interactions. Currently the focus in such efforts is on a particular exchange relationship within a single level of production. We consider the broader problem of assembling combinations of relationships across multiple levels to form complete, feasible supply chains.

In the example presented in Figure 1, Ford and GM need to acquire contracts for transmissions in order to produce particular models of cars. Ford can produce the transmissions in its own factories or acquire them from an independent transmission producer. GM currently does not have the capacity to produce the desired transmissions, and must outsource. The independent transmission producer has capacity to provide transmissions to either Ford or GM, but not both. Ford and the independent factory both require the services of a job shop for metal-working tasks, but the job shop does not have capacity to serve them simultaneously. Contracts with the job shop and with the two transmission factories are the scarce goods to be allocated.

---

2. http://www.covisint.com

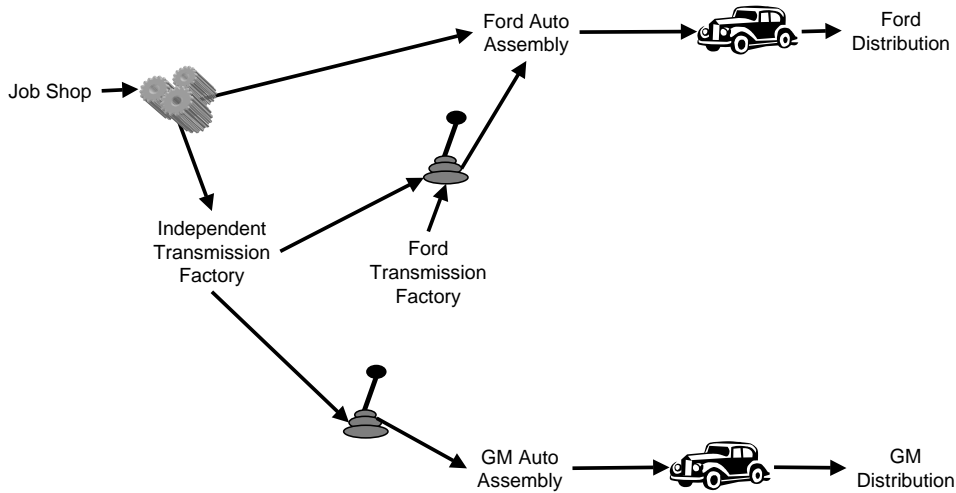

Figure 1: An automotive supply chain formation problem.

The limited capacity of the job shop entails certain constraints on feasible supply chains. Ford cannot acquire transmissions from the independent factory, because the job shop cannot serve the independent factory and Ford simultaneously. Additionally, Ford and GM cannot simultaneously be satisfied.

## 2.2 Problem Specification

We provide a formal description of the supply chain formation problem in terms of bipartite graphs. The two types of nodes represent goods and agents, respectively. A ***task dependency network*** is a directed, acyclic graph, $(V, E)$, with vertices $V = G \cup A$, where:

$$
\begin{aligned}
G &= \text{the set of goods,} \\
A &= C \cup \Pi, \text{the set of agents,} \\
C &= \text{the set of consumers,} \\
\Pi &= \text{the set of producers,}
\end{aligned}
$$

and a set of edges $E$ connecting agents with goods they can use or produce. There exists an edge $\langle g, a \rangle$ from $g \in G$ to $a \in A$ when agent $a$ can make use of one unit of $g$, and an edge $\langle a, g \rangle$ when $a$ can provide one unit of $g$. If an agent requires multiple units of a good as input, then we treat each unit as a separate edge, distinguishing them by subscripts. (Edges without explicit subscripts are interpreted as implicitly subscripted by "1".) For instance, if agent $a$ requires two units of $g$ as input, then its input edges are $\langle g, a \rangle_1$ and $\langle g, a \rangle_2$.

The various agent types are characterized by their position in the task dependency network. Each ***consumer***, $c \in C$, wishes to acquire one unit of one good from its set of consumable goods, $G_c \subseteq G$, where $\langle g, c \rangle \in E$ iff $g \in G_c$.

A ***producer*** can produce a single unit of an ***output*** good conditional on acquiring some ***input*** goods. With each producer $\pi \in \Pi$ we associate:

1. an input set, $I_\pi \subseteq G$, such that $g \in I_\pi$ iff there are edges $\langle g, \pi \rangle_k \in E$ for one or more $k$, and

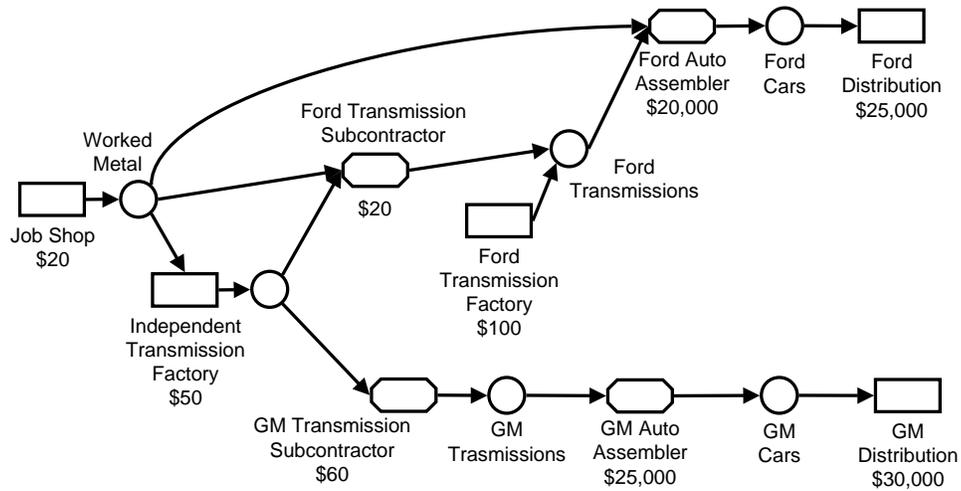

Figure 2: Network AUTO: A task dependency network for the automotive supply chain depicted in Figure 1.

2. a single output, $g_\pi \in G \setminus I_\pi$, such that $\langle \pi, g_\pi \rangle \in E$.

A producer's input goods are ***complementary*** in that the agent must acquire all of them in order to produce its output; it cannot accomplish anything with only a partial set. Alternate producers with the same output indicate different ways that a good can be produced.

Task dependency networks are constrained to be acyclic, that is, no agent produces goods that could be used to assemble its inputs through any chain of production. Although we might broadly view all global commerce as one large cycle of production and consumption, in practice, negotiations tend to be clustered within more limited scopes of concern, often referred to as "industries". The resulting supply chains are typically acyclic.

Figure 2 shows an example task dependency network for the automotive supply chain problem of Figure 1. Here the goods are indicated by circles, and agents by boxes. Producers with inputs are represented by curved boxes. The numbers under agent boxes represent production costs and consumption values, explained below. An arrow from an agent to a good indicates that the agent can provide that good, and an arrow from a good to an agent indicates that the agent can make use of the good. For instance, the producer labeled Ford Auto Assembly requires Worked Metal and Ford Transmissions in order to produce cars. Since the transmissions produced by the Ford Transmission Factory can be used only by Ford, we need to distinguish Ford and GM transmissions as separate goods. This in turn requires that we introduce Ford and GM Transmission Subcontractor producers to model the fact that the Independent Transmission Factory can be used to produce either type.

An ***allocation*** is a subgraph $(V', E') \subseteq (V, E)$. For $a \in A$ and $g \in G$, an edge $\langle a, g \rangle \in E'$ means that agent $a$ provides $g$, and $\langle g, a \rangle \in E'$ means $a$ acquires $g$. An allocation's vertices are the agents and goods incident on its edges:

1. An agent is in an allocation graph iff it acquires or provides a good:

   For $a \in A$, we have $a \in V'$ iff $\langle g, a \rangle \in E'$ or $\langle a, g \rangle \in E'$.

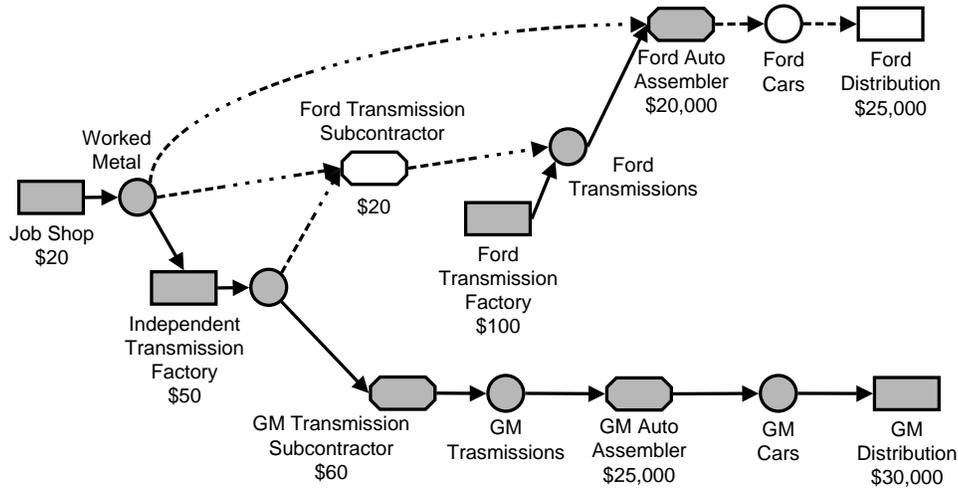

Figure 3: A solution to Network AUTO.

2. A good is in an allocation graph iff it is acquired or provided:

> For $g \in G$, we have $g \in V'$ iff $\langle g, a \rangle \in E'$ or $\langle a, g \rangle \in E'$.

A producer $\pi$ is **active** iff it provides its output. A **producer is feasible** iff it is inactive or acquires all its inputs. Consumers are always feasible.

Good $g$ is in **material balance** in $(V', E')$ iff the number of edges in equals the number out:

$$\left| \{(a,k) \mid \langle a,g \rangle_k \in E'\} \right| = \left| \{(a,k) \mid \langle g,a \rangle_k \in E'\} \right|.$$

An **allocation is feasible** iff all agents are feasible and all goods are in material balance. A **solution** is a feasible allocation that forms a partial ordering of feasible production, culminating in consumption. That is, some consumer acquires a good it desires:

> There exists a $\langle g, c \rangle \in E'$ such that $c \in C \cap V'$.

A solution may involve multiple consumers. If consumer $c$ is in a solution $(V', E')$ then we say that $(V', E')$ is a **solution for** $c$.

Figure 3 shows a solution allocation for the task dependency network of Figure 2. Shaded agents and solid arrows are part of the solution, with unshaded agents and dashed arrows indicating elements of the problem not part of the solution. Note that the Ford Auto Assembler wins an input, but is inactive. However, recall that inactive producers are feasible, hence the solution properties are met. We refer to the configuration of an inactive producer acquiring an input in an allocation as a **dead end**.

Each producer $\pi$ has some **production cost** $\kappa_\pi$ for providing a unit of its output. The cost might represent the value $\pi$ could obtain from engaging in some other activity (i.e., its opportunity cost), or some direct cost incurred in producing its output (but not including input costs). Since a producer provides at most one unit of one good, the total production cost to $\pi$, with output $g$, for allocation $E'$, is $\kappa_\pi$ if $\langle \pi, g \rangle \in E'$ and 0 otherwise.

We assume that a consumer has preferences over different possible goods, but wishes to obtain only a single unit of one good. Thus, a consumer $c$ obtains **value** $v_c(g)$ for obtaining a single unit of good $g$, and, for allocation $E'$, obtains value $v_c((V',E')) \equiv \max_{\langle g,c \rangle \in E'} v_c(g)$. In depicting task dependency networks, we display costs and values below the corresponding agent boxes.

**Definition 1 (value of an allocation)** *The* value of allocation $(V',E')$ *is:*

$$value((V',E')) \equiv \sum_{c \in C} v_c((V',E')) - \sum_{\pi \in \Pi} \kappa_\pi((V',E')).$$

**Definition 2 (efficient allocations)** *The set of* efficient allocations *contains all feasible allocations* $(V^*,E^*)$ *such that:*

$$value((V^*,E^*)) = \max_{(V',E') \subseteq (V,E)} \big(value((V',E')) \mid (V',E') \text{ is feasible}\big).$$

Task dependency networks describe the supply chain formation problem from a global perspective. In a decentralized approach to formation, we would generally not assume that an agent, or any other entity, has perfect or complete knowledge of the entire network. We generally do assume that all agents have perfect knowledge of their own costs, values, and goods of interest. When mediators facilitate the negotiations for goods (as in protocols described below), each agent knows of relevant mediators for its goods of interest. This knowledge includes all rules enforced by the mediators. Likewise, mediators know of the existence of all agents interested in their respective goods. Beyond that, a mediator knows only what the agents reveal through communication during negotiation. A mediator does not know the agents' true costs or valuations, nor is it aware of agents' preferences for goods outside of its direct scope of facilitation. We do not address in detail how agents and mediators achieve mutual awareness (i.e., how connections originate), but assume that it can be accomplished via some unspecified search, notification, or broadcast protocol.

## 3. Resource Contention

One natural candidate approach to supply chain formation is the CONTRACT NET protocol (Davis & Smith, 1983), the most widely studied algorithm for forming task performance relations among distributed agents. CONTRACT NET does indeed apply to our framework, as it employs local negotiation to achieve a hierarchical task decomposition. Although definitive characterization is difficult due to the many variants on CONTRACT NET in the literature (Baker, 1996; Davis & Smith, 1983; Dellarocas et al., 2000; Sandholm, 1993), it is fair to say that, generally, "request for quotes" proceed top down from the root task (right-to-left from consumers, in our network terminology), and contracting proceeds bottom-up (left-to-right towards consumers), selecting at each level among candidate "bids" received. (Variants of the protocol are primarily distinguished by the form of bids and selection criteria employed.) As a consequence, choices are made greedily, without reflecting ramifications upstream in the evolving chain.

This approach can form satisficing supply chains when there are sufficient resources to support the greedy selection. However, the basic CONTRACT NET protocol does not explicitly address resource scarcity or contention among multiple agents. Producers accept bids on inputs before it can be established whether this might cause infeasibility further upstream. Without lookahead or backtracking, CONTRACT NET might construct infeasible supply chains when there are limited resources.

For instance, a greedy protocol would not produce a solution for the network shown in Figure 4. Here, if all producers bid according to a common function monotone in cost, the output bid of producer a6 would be preferred to that of a5, because a6 can acquire its inputs cheaper. But since a7 must acquire the one available unit of good 4 to feasibly participate in a solution, a6 cannot be a part of the solution.

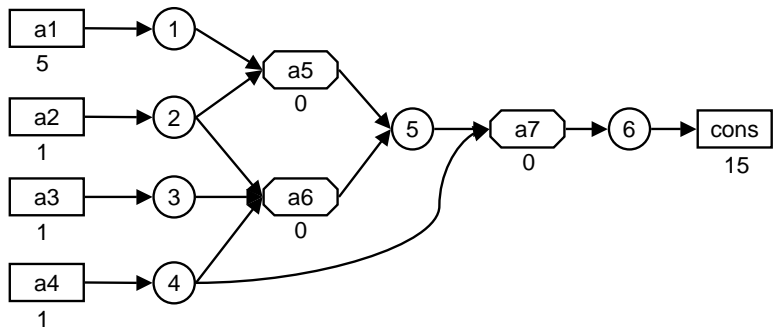

Figure 4: Network GREEDY-BAD: A network for which greedy protocols can produce infeasible allocations.

The issue of resource contention motivates our adoption of a market-based approach. The key idea is that prices can signal resource value and scarcity up and down the chain, enabling local decision making while avoiding the pitfalls of greedy one-pass selection or communication of global structure information.

## 4. Price Systems

A **price system** $p$ assigns to each good $g$, a nonnegative number $p(g)$ as its **price**. Prices are anonymous (i.e., not agent dependent) and linear in the quantity of goods. Intuitively, prices indicate the relative value of the goods, and agents use the prices to guide their local decision making.

We assume agents have **quasilinear** utility functions, defined by "money" holdings plus the value (or minus cost) associated with the allocation of goods. Agents wish to maximize their **surplus** with respect to prevailing prices.

**Definition 3 (surplus)** *The surplus, $\sigma(a, (V', E'), p)$, of agent $a$ with allocation $(V', E')$ at prices $p$, is given by:*

- $v_a((V', E')) - \sum_{\langle g,a \rangle \in E'} p(g)$, *if $a \in C$*

- $\sum_{\langle a,g \rangle \in E'} p(g) - \sum_{\langle g,a \rangle \in E'} p(g) - \kappa_\pi((V', E'))$, *if $a \in \Pi$.*

### 4.1 Price Equilibrium

Generally, an allocation $(V', E')$ is a **competitive equilibrium** at prices $p$ if $(V', E')$ is feasible and assigns to each agent an allocation that optimizes the agent's surplus at $p$. For our model, this means specifically:

- A producer's optimal choice is to be either active and feasible, or to acquire no goods. Hence, a producer in the allocation obtains nonnegative surplus by being active, and a producer not in the allocation would obtain nonpositive surplus by being active.

$$\forall \pi \in \Pi \cap V', \sum_{\langle \pi, g \rangle \in E} p(g) - \sum_{\langle g, \pi \rangle \in E} p(g) - \kappa_\pi \geq 0$$

$$\forall \pi \in \Pi \setminus V', \sum_{\langle \pi, g \rangle \in E} p(g) - \sum_{\langle g, \pi \rangle \in E} p(g) - \kappa_\pi \leq 0$$

- Because a consumer receives value for obtaining at most one good, a consumer's optimal choice is to obtain the good that gives it maximum nonnegative surplus, and to obtain no other goods at a positive price. Furthermore, a consumer not in the allocation (i.e., not obtaining any goods) would obtain nonpositive surplus from any good.

$$\forall c \in C \cap V', \exists \langle g, c \rangle \in E', \ g = \arg\max_{g' \in G} v_c(g') - p(g')$$

$$\wedge \ v_c(g) - p(g) \geq 0$$

$$\wedge \ \forall \langle g', E \rangle, \ g' \neq g, \ p(g') = 0$$

$$\forall c \in C \setminus V', \forall g \in G,$$

$$v_c(g) - p(g) \leq 0$$

Figure 5 shows an example of a competitive equilibrium for Network GREEDY-BAD. The prices are shown under their respective goods.

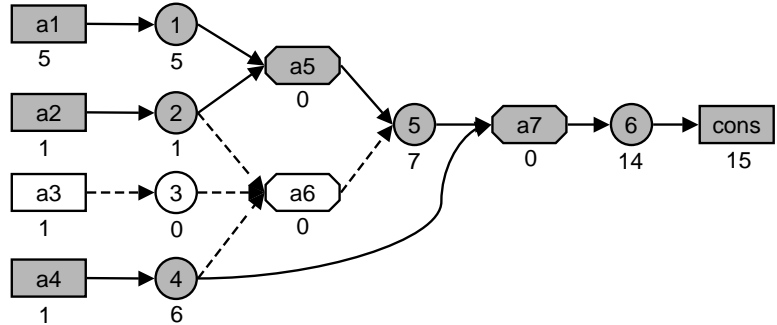

Figure 5: A competitive equilibrium for Network GREEDY-BAD.

A competitive equilibrium allocation is stable in the sense that no agent would want a different allocation at the equilibrium prices. Moreover, from equilibrium there is no way to reallocate the resources (including money transfers) so that some agent has greater surplus, without degrading some other agent's surplus. This absence of further gains from trade is referred to as **_Pareto optimality_**. Given quasilinear utility, price equilibria have been shown to be efficient under fairly general conditions (Bikhchandani & Mamer, 1997; Gul & Stacchetti, 1999; Ygge, 1998). This also holds for the particular case of task dependency networks, as stated in Corollary 4.

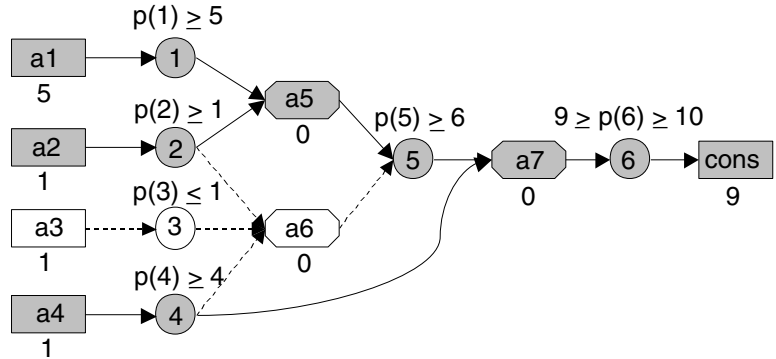

Figure 6: Network GREEDY-BAD with costs and values that do not support competitive equilibrium.

## 4.2 Existence of Competitive Equilibrium

Not all task dependency networks have competitive equilibria. Consider again Network GREEDY-BAD but with $v_{cons} = 9$, as shown in Figure 6. The allocation shown is the only efficient allocation, hence any equilibrium must support it. Recall that in equilibrium, active agents must obtain nonnegative surplus, and inactive producers must not be able to obtain positive surplus. The price inequalities under the goods follow from constraints on the surplus associated with agent activity. The lower bounds on the prices of goods 1, 2, and 5 ensure that producers a1, a2, and a5, respectively receive nonnegative surplus. The upper bound on 3 ensures that a3 could not obtain positive surplus. The lower bound on 4 ensures that a6 would receive nonpositive surplus. Propagating these bounds to 6, we see that $p(6) \geq 10$ to give a7 positive surplus, but also that $p(6) \leq 9$ to give cons nonnegative surplus. Since this is impossible, a competitive equilibrium cannot exist.

Technically, non-existence of equilibrium is due to complementarity of inputs for producers with discrete-quantity goods. In fact, complementarities are necessary to preclude competitive equilibrium in task dependency networks. A network has ***no input complementarities*** when all producers have at most one input.

**Theorem 1** *Competitive equilibria exist for any network with no input complementarities.*

We defer the proof of this and subsequent theorems to Appendix A.

Consider again Figure 6. The multiple undirected paths between 1 and 4 give rise to the lower bound on the price of good 6. It turns out that these undirected cycles are also necessary to preclude competitive equilibrium.

A ***polytree*** is a graph in which there is at most one undirected path from any vertex to another. Recall that in task dependency networks, if a producer uses multiple units of a good, then each unit is represented by a separate edge. It follows that an allocation is a polytree iff no more than one unit of a good is used to produce another given good, or used in multiple ways to produce a good.

**Theorem 2** *Competitive equilibria exist for any polytree.*

### 4.3 Approximate Price Equilibrium

We should generally expect that market protocols based on discrete price adjustments (such as the SAMP-SB protocol we describe in Section 5) would overshoot exact equilibria by at least a small amount. Therefore, our analysis emphasizes approximate equilibrium concepts (Demange et al., 1986; Wellman et al., 2001a). We introduce a particular type of approximation, $\lambda$-$\delta$-***competitive equilibrium***, defined in terms of parameters that bound the degree to which agents acquire suboptimal surplus. Intuitively, $\delta_b$ bounds the suboptimality of a consumer's surplus, $\delta_s$ bounds the suboptimality of a producer's surplus attributable to its output, and $\lambda_\pi^g$ bounds the suboptimality of a producer $\pi$'s surplus attributable to input $g$. As described in Section 5, these parameters also have special interpretation in our market protocol as applied to task dependency networks.

Denote as $H_a(p)$ the maximum surplus that agent $a$ can obtain in $(V, E)$, at prices $p$, subject to feasibility. That is,

$$H_a(p) \equiv \max_{(V',E') \subseteq (V,E)} \sigma(a, (V', E'), p)$$

such that $a$ is feasible at $(V', E')$.

**Definition 4 ($\lambda$-$\delta$-competitive equilibrium)** *Given the parameters:*

- $\delta_b$, $\delta_s \geq 0$,

- $\lambda_\pi^g$ *for all $\pi \in \Pi$ and all $g \in G$,*

*an allocation $(V', E')$ is in $\lambda$-$\delta$-competitive equilibrium at prices $p$ iff:*

1. *For all $a \in A$, $\sigma(a, (V', E'), p) \geq 0$.*

2. *For all $c \in C$, $\sigma(c, (V', E'), p) \geq H_c(p) - \delta_b$.*

3. *For all $\pi \in \Pi$, $\sigma(\pi, (V', E'), p) \geq H_\pi(p) - (\sum_{\langle g, \pi \rangle \in E} \lambda_\pi^g + \delta_s)$, and $\pi$ is feasible at $(V', E')$.*

4. *All goods are in material balance.*

Consider Network GREEDY-BAD with the same prices shown in Figure 5 except that $p(5) = 8$. This does not constitute an exact competitive equilibrium because a6, though inactive, could make a positive profit. However, if $\lambda_{a6}^2 + \lambda_{a6}^3 + \lambda_{a6}^4 + \delta_s \geq 1$, then since $H_{a6}(p) = 1$, a6 obeys Condition 3 and the allocation is a $\lambda$-$\delta$-competitive equilibrium at the specified prices.

**Theorem 3** *If $(V', E')$ is a $\lambda$-$\delta$-competitive equilibrium for $(V, E)$ at some prices $p$, then $(V', E')$ is a feasible allocation with a nonnegative value that differs from the value of an efficient allocation by at most $\sum_{\pi \in \Pi} [\sum_{\langle g, \pi \rangle \in E} \lambda_\pi^g + \delta_s] + |C|\delta_b$.*

A $\lambda$-$\delta$-competitive equilibrium corresponds to the standard notion of competitive equilibrium when $\delta_b = \delta_s = 0$, and $\lambda_\pi^g = 0$ for all $\pi$ and $g$.

**Corollary 4 (to Theorem 3)** *A competitive equilibrium allocation is efficient.*

As noted in Section 4.1, this is consistent with previously established results.

## 4.4 Valid Solutions

In the following sections we show that λ-δ-competitive equilibria can be a useful concept for analyzing decentralized market protocols. However, such protocols do not always reach λ-δ-competitive equilibria for all networks. Hence we also consider weaker constraints on prices, consistent with a lesser degree of agent optimization in a solution allocation.

We say that a solution $(V', E')$ is **valid** with respect to prices $p$ if:

1. Each consumer in the solution pays no more than its value for a single good. That is, for all $c \in C \cap V'$, there exists a single $\langle g, c \rangle \in E'$ such that

$$p(g) \leq v_c(g),$$

   and $p(g') = 0$ for all $g' \neq g$ such that $\langle g', c \rangle \in E'$.

2. None of the active producers are unprofitable. For all $\pi \in \Pi \cap V'$ where $\langle \pi, g_\pi \rangle \in E'$ we have $\sigma(\pi, (V', E'), p) \geq 0$. Note that solution validity does not preclude an *inactive* producer from being unprofitable (i.e., it admits dead ends).

Note that (1) effectively states that consumers do not obtain negative utility, which is weaker than the competitive equilibrium conditions in that it does not require consumers to receive their optimal allocation. Similarly, (2) does not require producers to optimize, as in competitive equilibrium, but only requires nonnegative utility for *active* producers.

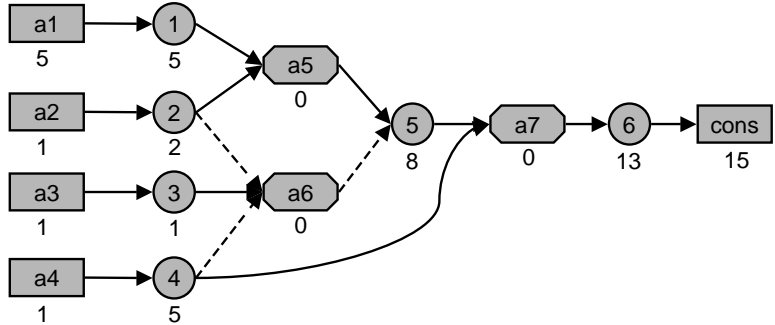

Figure 7: A valid solution for Network GREEDY-BAD.

Figure 7 shows an example valid solution, with the same underlying costs and values as in Figure 5. Because it allows dead ends, validity does not directly provide useful bounds on the inefficiency of an allocation.

## 5. SAMP-SB Protocol

The preceding section introduces some static properties of price configurations and allocations. Here we address the problem of how prices might be obtained. To compute prices and allocations, we must elicit information bearing on the relative value of goods, through some systematic communication process. Mechanisms that determine market-based exchanges based on messages from agents are called **auctions** (McAfee & McMillan, 1987).

The agents' ***bidding policies*** represent their strategies for interacting with the auctions. Whereas the auction mechanism may be designed by a central authority, bidding policies are generally determined by individual agents. To understand the implications of the auction design requires an analysis of the ***market protocol*** that arises from the combination the auction mechanism and the agent bidding policies.

The space of potential auctions is expansive (Wurman et al., 2001), and definitive theoretical results are currently known only for fairly limited classes of problems (Bikhchandani & Mamer, 1997; Demange & Gale, 1985; Gul & Stacchetti, 2000; Klemperer, 1999; McAfee & McMillan, 1987). Complementarities with discrete goods, which can cause nonexistence of price equilibria, also greatly complicate auction design and analysis of auctions (Milgrom, 2000).

For our supply-chain domain, we have investigated a particular protocol, called ***SAMP-SB*** (Simultaneous Ascending (M+1)st Price with Simple Bidding). As demonstrated below, SAMP-SB can produce good allocations which, in some cases, are consistent with competitive price equilibrium theory.

## 5.1 Auction Mechanism

The SAMP-SB mechanism comprises a set of auctions, one for each good. Auctions run simultaneously, asynchronously, and independently, without direct coordination. Agents interact with the auctions by submitting bids for goods they wish to buy or sell. A bid is of the form: $((q_1 \ p_1) \ldots (q_n \ p_n))$. Each pair $(q_i \ p_i)$ indicates an offer to buy or sell the good, with $q_i$ indicating the quantity of the offer and $p_i$ indicating the price. If $q_i > 0$, it is an offer to buy $q_i$ units of the good for no more than $p_i$ per unit, and we refer to it as a ***buy offer***. If $q_i < 0$, it is an offer to sell $q_i$ units for no less than $p_i$ per unit, and we refer to it as a ***sell offer***. Because no agent both buys and sells the same good in a task dependency network, a bid contains either all positive or all negative quantity offers. Bids possess what is sometimes called "additive-OR" semantics—the offers are treated exactly as if they came from separate bids, hence the auction can match any of the individual offers independently. Without loss of generality, we henceforth impose the restriction $|q_i| = 1$ for all offers in all bids, continuing to allow that agents may submit multiple offers in a bid.

When an auction receives a new bid, it sends each of its bidders a ***price quote*** specifying the price that would result if the auction ended in the current bid state. Price quotes are not issued until all initial bids are received, but are subsequently issued immediately on receipt of new bids. Because some offers may be tied at the current price, this information alone is not sufficient for an agent to tell whether it is winning an offer placed at that price. To clarify this ambiguity, the price quote also reports to each bidder the quantity it would buy or sell in the current state. The same prices are sent to all bidders, but the reported winning state is specific to the recipient. Agents may then choose to revise their bids in response to the notifications (if an agent does not wish to change its bid, inaction leaves its previous bid standing in the auction).

We assume that communication is reliable but asynchronous.[3] That is, all messages sent eventually reach their recipients, although we impose no bound on the delays. Agents and auctions use message IDs to ensure that they handle messages in the appropriate order. Note that even if all auctions and agents have deterministic behaviors, an overall run of SAMP-SB may be nondeterministic due to this asynchrony.

---

3. Technically, we adopt the model of *asynchronous reliable message passing* systems (Fagin et al., 1995).

Under asynchrony, it is helpful for the auction to send the ID of the most recent bid received from the agent with its price quote. An agent responds only to a price quote that reflects its most recent bid sent. Without this device, an agent can have difficulty establishing feasibility, as its understanding of its input and output bid states may be based on nonuniformly delayed reports.

Bidding continues until **quiescence**, a state where all messages have been received, no agent chooses to revise its bids, and no auction changes its prices, ask prices, or allocation. At this point, the auctions **clear**; each bidder is notified of the final prices and how many units it transacts in each good. Note that a quiescent system is not necessarily in a solution state or (approximate) equilibrium state.

Although detecting quiescence is straightforward in a centralized system, in a decentralized, asynchronous system we need to perform the operation using only local message passing. In previous work (Wellman & Walsh, 2000), we described a protocol for detecting quiescence in general distributed negotiations, based on a well-known termination-detection algorithm.

Each auction runs according to (M+1)st-price rules (Satterthwaite & Williams, 1989, 1993; Wurman et al., 1998). The (M+1)st price auction is a variant of the (second-price) Vickrey auction (Vickrey, 1961), generalized to allow for the exchange of multiple units of a good. Given a set of offers including $M$ units offered for sale, the (M+1)st-price auction sets a price equal to the price of the (M+1)st highest offer over *all* of the offers. The price can be said to separate the winners from the losers, in that the winners include all sell offers strictly below the price and all buy offers strictly above the price. Some agents that offer at the (M+1)st price also win; in case of ties, offers submitted earlier have precedence. Winning buy and sell offers are matched one-to-one, and pay (or get paid) the (M+1)st price.

When issuing price quotes, the auction reports both the price (i.e., the current going price, or (M+1)st price), $p(g)$ and the **ask price**, $\alpha(g)$ of the good $g$. The ask price specifies the amount above which a buyer would have to offer in order to buy the good, given the current set of offers. The ask price is determined by the price of the $M$th highest of all offers in the auction, hence $\alpha(g) \geq p(g)$. For instance, if we have buy bids 12, 10, and 6 and sell bids 15, 11, and 8, $p(g) = 10$, $\alpha(g) = 11$, and if the auction is in quiescence, the buy bids 12 and 10 would match the sell bids 15 and 11 and trade at $p(g) = 10$.

Because a producer has complementary inputs, ensuring feasibility is a challenging problem, requiring careful design. The auctions run simultaneously, and each auction requires that the prices of an agent's successive buy offers increase by no less than some (generally small) positive number $\delta_b$ and the prices of successive sell offers increase by no less than $\delta_s$.[4] An auction can enforce the ascending rule by simply rejecting an agent's offer if the price does not increase by $\delta_b$ or $\delta_s$. By constraining the direction of price changes, this design gives producers a more accurate indication of the relative prices for inputs and outputs than if prices were allowed to fluctuate in both directions.

The ascending bid restriction ensures ascending auction prices, with one technicality. Due to asynchrony and immediate issuance of price quotes, if the initial bid from an agent arrives after a higher bid, the price quote could decrease. This can be handled simply at the auction by issuing no price quotes until some specified period of time after the auction opens. After the first price quote is issued, the auction accepts new bids only from agents that had previously placed bids.

It is common in auction literature and practice to place an ascending restriction on buy-offer prices. It may seem counterintuitive—and is in fact atypical—to place the same restriction on sell-offer prices. However, such an ascending offer price restriction ensures that price quotes rise monotonically as the auctions progress. Section 5.4 shows how an ascending–offer-price restriction for both buy and sell offers serves a key role in establishing the relationships between system quiescence and solution convergence of the system.

---

4. These rules differ from those of a more typical simultaneous ascending auction (Demange et al., 1986; Milgrom, 2000), which specify that agents must submit offer prices that are at least an increment above the current price.

## 5.2 Bidding Policies

Although designers of negotiation mechanisms do not generally have control over the agents' behaviors, any conclusions about the outcome of a mechanism must be based on some assumptions about these behaviors. A typical assumption in economics is that agents are rational in some sense, for example that they play policies that form a Bayes-Nash equilibrium. However, as discussed in Section 7.1, the complexity of supply chain formation markets is beyond the current state-of-the-art in analyzing Bayes-Nash equilibria with simultaneous ascending auctions. Instead, our analysis assumes that the agents follow a simple, non-strategic bidding policy, described in this section. Other variations may be reasonable, or perhaps better in some respects than the policies we describe. Rather than explore the range of possibilities, we chose in this work to investigate a particular set of policies in depth. Our chosen policies obey the ascending offer restriction enforced by the auction, respect the locality of information in that they require no knowledge of other agents in the system, and are myopic in that they use only information provided by the current price quotes, without forecasting future prices.

Recall that a consumer wishes to acquire a single good that maximizes its surplus at the given prices. We assume that a consumer initially offers zero for each good of interest. So long as it is winning a good, it does not change its offer. Whenever it is not winning a good, it offers $p(g^*) + \delta_b$ for good $g^* = \arg\max_{g \in G}(v_c(g) - p(g) - \delta_b)$ if $v_c(g^*) - p(g^*) - \delta_b \geq 0$, otherwise it stops bidding.

A producer's objective is much more complex, namely to maximize the difference between the price it receives for its output and the total price it pays for its inputs, while remaining feasible. We assume that a producer initially offers zero for each of its input goods, and gradually increases these offers to ensure feasibility. It raises its offer price for an input good by $\delta_b$ if and only if the price quotes indicate that it is losing that good but winning its output.

We assume that producer $\pi$ bids for its output good $g_\pi$ in an effort to recover its production cost and the **_perceived costs_** of its inputs. The producer places its first output offer only after receiving the first price quotes for all its inputs, and subsequently updates its output offer whenever it receives a new price quote on any input. For simplicity, consider the case in which $\pi$ has one offer (each at quantity one) for each input. If $\pi$ is currently winning an input $g$, its perceived cost, $\hat{p}_\pi(g)$ of $g$ is simply $p(g)$. When $\pi$ is not currently winning $g$ with a particular offer, $\hat{p}_\pi(g) = \max(\alpha(g), p(g) + \delta_b)$. If $\beta$ is the price of the previous offer made by $\pi$ for $g_\pi$, then when its perceived costs increase, $\pi$ offers $\max(\beta + \delta_s, \sum_{\langle g,\pi \rangle \in E} \hat{p}_\pi(g))$ for its output $g_\pi$. If $\pi$ has multiple offers for a good $g$, then it assumes a separate perceived cost with respect to each offer, and bids for its output accordingly. Figure 8 shows how a producer would bid next as a function of the current prices and its current offers, when $\delta_b = 1$ and $\delta_s \leq 2$.

Note that throughout the negotiation, a producer places bids for its output goods before it has received commitments on its input goods. Producers counteract potential risk by continually updating their bids based on price changes and feasibility status. A producer reduces exposure to dead ends by incrementing its offer prices on inputs by minimal amounts and only when necessary.

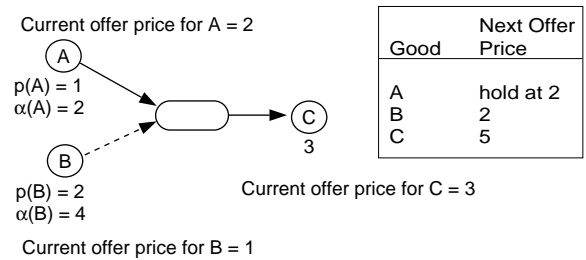

Figure 8: A producer's next offers, according to SAMP-SB, when $\delta_b = 1$ and $\delta_s \le 2$. The dashed arrow from good B indicates that the producer is currently losing B. The solid arrows indicate that the producer is currently winning goods A and C.

### 5.3 Bidding with General Preferences

The task dependency network model represents fairly simple production capabilities and consumer utility. Here we discuss some natural potential extensions of the bidding policies to a broader class of capabilities and preferences.

A producer capable of variable-unit production could bid exactly as if it were multiple identical producers. Such a producer would maintain separate offers in its bids for each unit, and update the separate offers independently. Similarly, a consumer with additive value for multiple goods, or multiple units of a good, could bid for each unit of each good as if it were a separate consumer.

A producer with alternatives on some input, independent of other inputs, can switch its bidding to the currently cheapest option. Subtle issues can arise for a producer that has alternative input *sets*, particularly when it is tentatively winning parts of the sets. One option would be to focus bidding on the set with the lowest perceived cost, which may include a premium for goods not in the tentatively winning set. Alternatively, the producer could assume that it will definitely win its tentatively won goods and effectively treat them as sunk costs. Fractional accounting of sunk costs may also be reasonable. Similar considerations arise for extensions presenting complex consumption choices.

### 5.4 Properties of SAMP-SB

In this section we describe a number of theoretical properties of SAMP-SB. In Section 5.4.1 we describe properties relating to convergence to quiescence, in Section 5.4.2 we present properties relating to efficiency and convergence to price equilibrium, and in Section 5.4.3 we present properties relating to solution convergence.

#### 5.4.1 CONVERGENCE TO QUIESCENCE

The SAMP-SB auctions and bidding policies guarantee that the system will always reach quiescence.

**Theorem 5** *SAMP-SB reaches quiescence after a finite number of bids have been placed.*

However, convergence can take a long time.

**Observation 6** *In an asynchronous environment, it is possible that a run of the protocol may require a number of bids that is exponential in the network size, and not a function of the consumer value.*

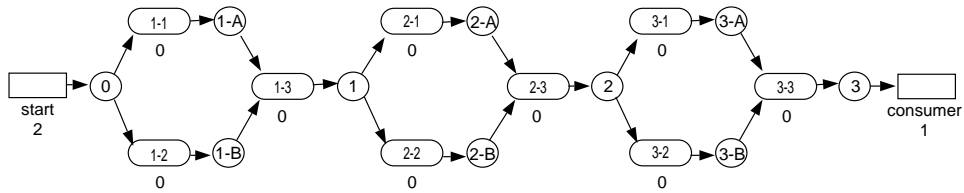

Figure 9: Network EXPONENTIAL: A network that may require an exponential number of bids to reach quiescence.

Figure 9 shows Network EXPONENTIAL, which illustrates this observation. The agent named start places a one-time bid to sell one unit of good 0 for $2. Since $\alpha(0) = 2$, and producers 1-1 and 1-2 are initially losing their input bids, these agents each offer a price of 2 for their output goods. Producer 1-3 will receive the new price quotes for goods 1-A and 1-B asynchronously, hence may update its bid for good 1 twice, offering a price of 2 the first time and a price of 4 the second time. Continuing with this process, we see that producer 3-3 updates its bid for good 3 up to eight times. If we extend this network and maintain labeling consistent with Figure 9, then producer n-3 would place $O(2^n)$ bids for good n. Note however, that if bids and price quotes are propagated synchronously, the exponential growth would not occur.

In the example above, most of the bids are actually superfluous in that they do not meaningfully affect the outcome of the protocol. This appears often true of situations exhibiting the worst-case behavior described. To capture the distinction between relevant and irrelevant bidding, we introduce the notion of ***quasi-quiescence***, a persistent state from which all subsequent bids effectively do not matter for solution convergence. SAMP-SB convergence to quasi-quiescence requires a number of "meaningful" bids that can be bounded by the size of the network and the value of the maximum consumer value.

**Definition 5 (quasi-quiescent)** *A run of SAMP-SB is in a* quasi-quiescent *state when, for any consumer or active producer* τ, *all bids by* τ *have been received and* τ *would not change its bids in response to any price quotes already received or transmitted by auctions.*

Clearly, the requirements of quasi-quiescence are subset of the requirements for quiescence.

**Observation 7** *A quiescent state is a quasi-quiescent state.*

**Theorem 8** *If a run of SAMP-SB reaches a quasi-quiescent state, then it remains in a quasi-quiescent state. Furthermore, neither the allocation nor the prices p subsequently change.*

This theorem means that, once quasi-quiescence is reached, all subsequent bids effectively do not matter in terms of equilibrium and solution convergence.

**Corollary 9 (to Theorem 8)** *The quiescent state of SAMP-SB is a* λ-δ-*equilibrium or valid solution iff the first quasi-quiescent state reached is a* λ-δ-*equilibrium or valid solution, respectively.*

The following theorem establishes a bound on the number of relevant bids necessary to reach quasi-quiescence.

**Theorem 10** *SAMP-SB reaches a quasi-quiescent state after a number of bids bounded by a polynomial of the size of the network and the value of the maximum consumer value have been placed by consumers and active producers.*

Our previously mentioned quiescence-detection protocol (Wellman & Walsh, 2000) can also detect quasi-quiescence, and thus terminate negotiations when it is reached.

### 5.4.2 EFFICIENCY AND CONVERGENCE TO PRICE EQUILIBRIUM

We intentionally use $\delta_b$ and $\delta_s$, to parametrize both SAMP-SB and our concept of $\lambda$-$\delta$-competitive equilibrium. With an interpretation of $\lambda_\pi^g$ in terms of prices and ask prices, we can specify necessary and sufficient conditions for which the *result* of SAMP-SB corresponds to a $\lambda$-$\delta$-competitive equilibrium.

**Theorem 11** *The prices and allocation determined in quiescence by the SAMP-SB protocol is a $\lambda$-$\delta$-competitive equilibrium, with $\lambda_\pi^g = \max(\alpha(g) - p(g), \delta_b)$, iff no inactive producer buys any positive-price input.*

From Theorems 3 and 11, we can establish bounds on the inefficiency of a $\lambda$-$\delta$-competitive equilibrium, parametrized by $\lambda_\pi^g = \max(\alpha(g) - p(g), \delta_b)$ for each good. In some cases, the difference between $\alpha(g)$ and $p(g)$ may be quite high. However, we can actually establish a tighter bound.

**Theorem 12** *If $(V', E')$ is a $\lambda$-$\delta$-competitive equilibrium computed by SAMP-SB, then $(V', E')$ has a nonnegative value that differs from the value of an efficient allocation by at most $\sum_{\pi \in \Pi}(|\{\langle g, \pi \rangle \in E\}|\ \delta_b + \delta_s) + |C|\delta_b$.*

Note that the theorem replaces $\lambda_\pi^g$ from Theorem 3 with $\delta_b$ in the bound.

A network is a ***tree*** if it is a polytree with no more than one consumer.

**Theorem 13** *The quiescent state of SAMP-SB is a $\lambda$-$\delta$-competitive equilibrium for a tree.*

We are unaware of other general network structures for which SAMP-SB is guaranteed to converge to a $\lambda$-$\delta$-competitive equilibrium. However, Theorem 11 implies that we can improve allocations if we modify SAMP-SB to avoid dead ends. We say that a bidding policy is ***safe*** for a producer if the producer cannot obtain a negative surplus in quiescence. It is clear that if a protocol is safe for *all* producers, then it will converge to $\lambda$-$\delta$-competitive equilibrium.

In SAMP-SB we have assumed that a producer updates buy and sell offers simultaneously in response to price quotes. This policy is not safe, even for single-input producers, because the producer bids for its input based on the state of its *standing* offer for its output, rather than the offer it is about to place. The producer would get negative surplus if it does not win its new output offer but gets stuck winning its new input offer. However, a slight variant of the bidding policy, which we call ***safe SAMP-SB***, is safe for any single-input producer. With this protocol, a producer updates its input bids only when it would not update, and it currently winning, its most recent output offer. Clearly, safe SAMP-SB has the same static properties as SAMP-SB, hence Theorem 12 applies to safe SAMP-SB.

**Theorem 14** *The quiescent state of safe SAMP-SB is a $\lambda$-$\delta$-competitive equilibrium for a network with no input complementarities.*

Safe SAMP-SB is not guaranteed to be safe for producers with multiple inputs in arbitrary networks, nor do we know of any safe producer bidding policy that ensures safety for producers in any arbitrary network (other than degenerate policies such as not bidding).

Safe SAMP-SB may take longer to reach quiescence than regular SAMP-SB. With safe SAMP-SB, a producer must always wait for notification of the results of pending output offers before increasing input offers. For a producer to win an output offer may require propagations of many messages through various paths in the network before buyers of the output good would increase their buy offer prices for that good. The resulting delay would be greater than the local delay in communicating with the output good auction.

That non-$\lambda$-$\delta$-competitive equilibrium runs of SAMP-SB result in dead ends suggests a potential source of significant efficiency loss. For example, Figure 7 shows the result of a run of SAMP-SB on Network GREEDY-BAD. This valid solution has a dead end at producer a6. Since producer a3 incurs its cost of \$1 to provide good 3 to a6, but does not contribute to any value in the system, this dead end is pure waste from a global efficiency perspective. The allocation is undesirable directly for producer a6 because it is committed to pay \$1 for an input it cannot use. With large networks or costs, dead ends can result in significant efficiency losses and negative profits to individual agents.

We propose a ***contract decommitment protocol*** to remove dead ends after SAMP-SB reaches quiescence. According to the decommitment protocol, each inactive producer can decommit from its contracts for its inputs for which it would pay a positive price. The protocol is applied recursively to the producers that lose their outputs due to decommitment. When the decommitment process terminates, agents exchange goods as specified by the remaining contracts. We refer to SAMP-SB with decommitment as ***SAMP-SB-D***.

In Figure 7, producer a6 would decommit from its contract with a3. Clearly, Theorem 11 implies that no agent decommits iff SAMP-SB produced a $\lambda$-$\delta$-competitive equilibrium. Moreover, if we remove from consideration all producers that decommit, the remaining agents are in $\lambda$-$\delta$-competitive equilibrium.

Decommitment has the benefit that, whereas some producers can lose money in the SAMP-SB protocol, no agent receives a negative surplus from participating in SAMP-SB-D. However, this is achieved by making the auction allocations non-binding, which is undesirable to the producers who lose their output sales to decommitments. It also begs the question of how to enforce the requirement that inactive producers be the only agents that decommit.

In addition to dead ends, efficiency can also be lost if SAMP-SB fails to find a solution when a positive value solution exists, or if SAMP-SB forms a solution with value inferior to an efficient solution (dead ends are not necessarily mutually exclusive of these two cases). In Section 6 we describe an experimental analysis of the efficiency, the source of inefficiency, and equilibrium attainment of SAMP-SB in a set of networks.

### 5.4.3 SOLUTION CONVERGENCE

Recall that SAMP-SB always converges to a valid solution (specifically a $\lambda$-$\delta$-competitive equilibrium) for networks with tree structures, and the safe variant converges for networks with no input complementarities. The following theorem shows that, with sufficiently high consumer value, regular SAMP-SB can always converge to a (possibly non-equilibrium) valid solution for polytrees.

**Theorem 15** *If $(V,E)$ is a polytree with a solution that assigns good g to consumer c, then given all other costs and values, there exists a value $v_c(g)$ such that SAMP-SB is guaranteed to converge to a valid solution $(V',E')$ for c.*

Because dead ends may result, we cannot usefully bound the inefficiency of the solution reached by SAMP-SB in a polytree.

For general network structures, the prices of all sell offers for all consumers' goods could rise above their values, in which case the system will necessarily reach quasi-quiescence in a non-solution state. If, however, quasi-quiescence is reached before the price of some consumer's good reaches its value for the good, we have a valid solution.

**Theorem 16** *If SAMP-SB reaches quasi-quiescence with $p(g) < v_c(g)$ for some $\langle g,c \rangle \in E$, $c \in C$, then the system's state represents a valid solution.*

The next theorem establishes conditions under which a valid solution state will immediately lead to quasi-quiescence.

**Theorem 17** *If a run of SAMP-SB in $(V,E)$ is in a valid solution state such that:*

- *each consumer c is either winning an offer or $p(g) + \delta_b > v_c(g)$ for all $\langle g,c \rangle \in E$,*

- *all agents have correct beliefs about which goods they are currently winning,*

- *all bids from consumers and active producers have been received in response to the current price quotes,*

- *and no sell offers are lost due to tie breaking,*

*then after the subsequent price quote from each auction, the system will be in a quasi-quiescent state with a valid solution.*

Although SAMP-SB is not guaranteed to converge to a solution, the fact that the problem of finding a solution is NP-Complete (Walsh et al., 2003) should lead us to expect that there are problems for which SAMP-SB would converge to a solution only after an exponential number of meaningful bids. Since the number of meaningful bids is bounded by a polynomial of the maximum consumer value, we should further expect that there exist networks for which SAMP-SB can converge to a solution only with a exponential consumer values. In practice we find that we can construct problems for which the consumer value must be exponential in order for SAMP-SB to converge to a solution (Walsh et al., 2003). However, we have run many simulations for which the required value is much more reasonable (Walsh et al., 2003).

For some networks, costs, and values, SAMP-SB *cannot* converge to a valid solution with some values of $\delta_b$ and $\delta_s$, no matter how high the consumer value. One example (the simplest we have been able to construct) is Network NO-CONVERGE, shown in Figure 10. Observe that a solution must include agent a8, but cannot include a7. Agent a6 always offers a price of at least $p(2) + 20$ for good 4, hence a8 cannot win two units of good 4 for less than $p(2) + 20$ each. Thus agent a8 will always offer a price of at least $2p(2) + 40$ for good 5. Since agent a7 will never offer a price more than $2p(2) + 2\lambda_{a7}^2$ for good 5, agent a8 could only win good 5 if $\lambda_{a7}^2 \geq 20$. But, for this to occur, we must have $\delta_b \geq 20$. A more thorough analysis, taking into account the dynamics of SAMP-SB, shows we must have $\delta_b \geq 40$ and $\delta_s = 0$ to obtain a valid solution in quiescence, and then only for certain patterns of asynchrony.

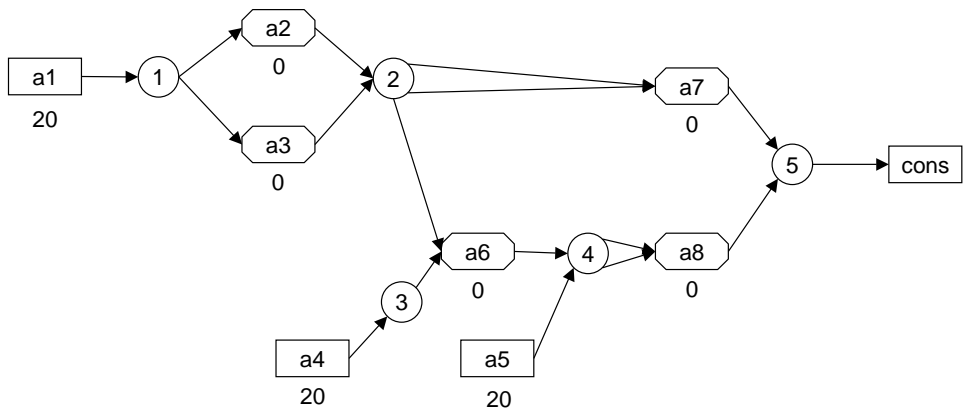

Figure 10: Network NO-CONVERGE: A network for which SAMP-SB cannot converge to a solution for certain values of $\delta_b$ and $\delta_s$.

## 6. Empirical Performance of SAMP-SB

Whereas our analytic results provide some insight into SAMP-SB and its variants, they do not support a comprehensive characterization of performance, except for certain special-case network structures. In order to gain further understanding of the effectiveness of SAMP-SB and SAMP-SB-D, we performed an empirical study based on protocol simulations on sample task dependency networks.

### 6.1 Setup

Our investigation focuses on a small set of networks exhibiting a variety of structural properties: SIMPLE (Figure 11), UNBALANCED (Figure 12), TWO-CONS (Figure 13), BIGGER (Figure 14), and MANY-CONS (Figure 15). We also also studied Network GREEDY-BAD (Figure 4).

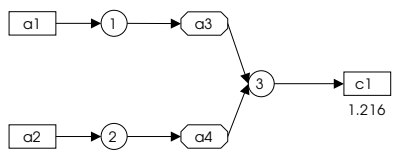

Figure 11: Network SIMPLE.

We ran experiments on multiple instances of each network. For each instance we randomly chose producer costs uniformly from $[0, 1]$, but for each consumer in a network, we calculated a fixed value so that, excluding all other consumers, there exists a positive-surplus solution for this consumer with 0.9 probability. We determined consumer values via simulation, assuming the specified distributions of producer costs. We discarded all instances whose efficient solutions had value zero. We set $\delta_b = \delta_s = .01$.

To test the effect of competitive equilibrium existence on the performance of the protocols, we generated instances of UNBALANCED, TWO-CONS, and GREEDY-BAD with costs that admit competitive equilibrium and with costs that do not. Because SIMPLE and MANY-CONS are polytrees, we know from Theorem 2 that all instances thereof have competitive equilibria. We were not able to generate no-equilibrium instances of BIGGER with the given cost distributions.

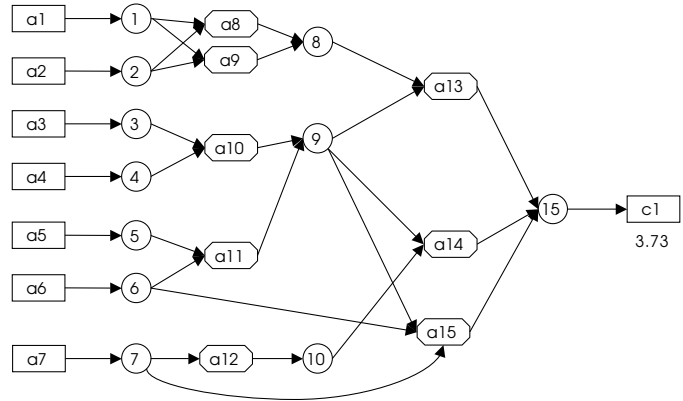

Figure 12: Network UNBALANCED.

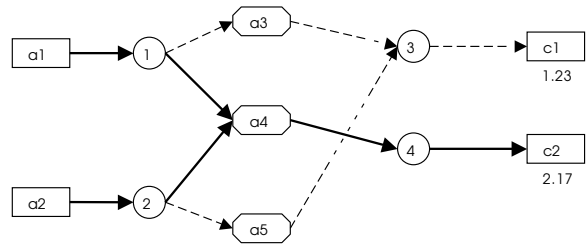

Figure 13: Network TWO-CONS.

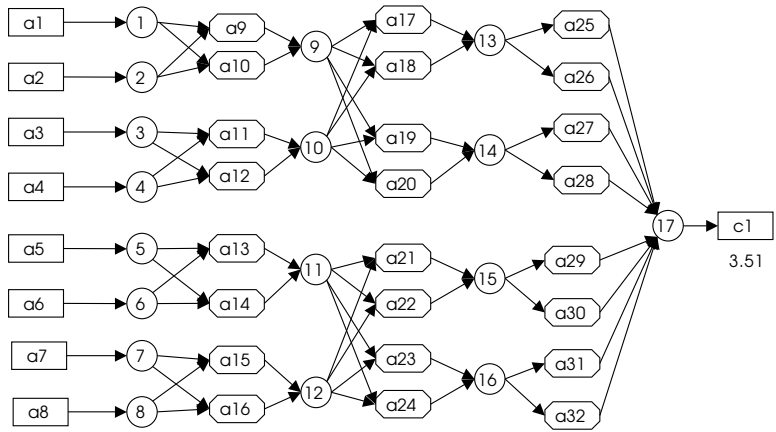

Figure 14: Network BIGGER.

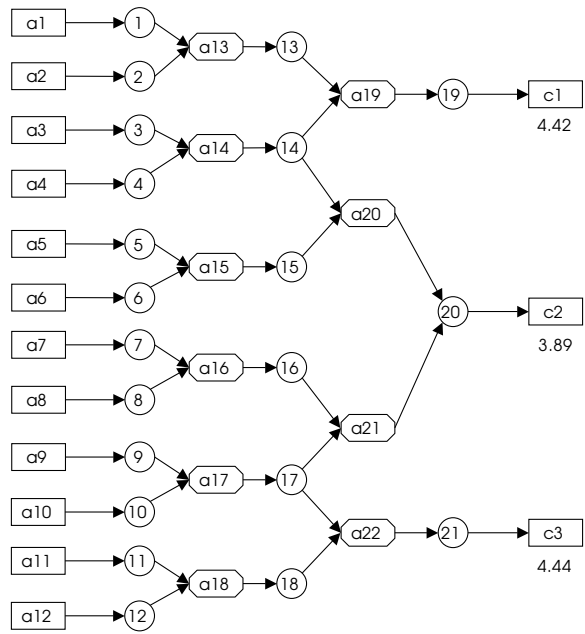

Figure 15: Network MANY-CONS.

To generate an instance with a desired type of cost structure (equilibrium or no-equilibrium) we repeatedly chose sets of producer costs randomly from the uniform distribution until the desired property was met. In the experiments, we determined whether competitive equilibrium existed—given complete information about the network structure, values, and costs—using the following procedure. Recall that a competitive equilibrium is always efficient (Corollary 4). Hence, given an optimal allocation $(V^*, E^*)$, we attempt to solve the system of linear equations that characterize a competitive equilibrium, as described in Section 4.1. If a solution to the equations exists, the resulting prices constitute a competitive equilibrium, otherwise no equilibrium exists. We used CPLEX, a commercial mixed-integer-linear programming package, to find the efficient allocation and to solve the corresponding equilibrium equations.

For each type of cost structure in each network, we tested 100 random instances, with the exception of SIMPLE, for which we tested 3220 instances.[5] For each instance and each protocol, we measured the **efficiency**—the fraction of the efficient value—attained by SAMP-SB and SAMP-SB-D. We also measured the percentage of available surplus (i.e., percentage of the value of an optimal solutions) obtained by the producers.

## 6.2 Results

We classify the efficiency of a run of the protocols in one of four ways: Negative, Zero, Suboptimal (but positive), and Optimal efficiency. Table 1 shows the distribution of the efficiency classes in our experiments. Note that SAMP-SB-D cannot produce negative efficiency, by construction.

---

5. We tested more instances of SIMPLE as a part of a broader study (Walsh et al., 2000).

| Network | SAMP-SB % of instances | | | | SAMP-SB-D % of instances | | |
|---|---|---|---|---|---|---|---|
| | Neg | Zero | Sub | Opt | Zero | Sub | Opt |
| SIMPLE | 0.0 | 0.3 | 0.0 | 99.7 | 0.3 | 0.0 | 99.7 |
| UNBALANCED, case: | | | | | | | |
| – equilibrium exists | 5.0 | 1.0 | 7.0 | 87.0 | 1.0 | 1.0 | 98.0 |
| – no equilibrium exists | 100.0 | 0.0 | 0.0 | 0.0 | 100.0 | 0.0 | 0.0 |
| TWO-CONS, case: | | | | | | | |
| – equilibrium exists | 11.0 | 0.0 | 6.0 | 83.0 | 0.0 | 3.0 | 97.0 |
| – no equilibrium exists | 18.0 | 0.0 | 78.0 | 4.0 | 1.0 | 95.0 | 4.0 |
| BIGGER | 0.0 | 0.0 | 4.0 | 96.0 | 0.0 | 0.0 | 100.0 |
| MANY-CONS | 27.0 | 0.0 | 56.0 | 17.0 | 0.0 | 2.0 | 98.0 |
| GREEDY-BAD, case: | | | | | | | |
| – equilibrium exists | 4.0 | 0.0 | 21.0 | 75.0 | 1.0 | 0.0 | 99.0 |
| – no equilibrium exists | 100.0 | 0.0 | 0.0 | 0.0 | 100.0 | 0.0 | 0.0 |

Table 1: Distribution of efficiency classes from SAMP-SB and SAMP-SB-D. Efficiency classes: Negative (Neg), Zero, Suboptimal (Sub), and Optimal (Opt).

Recall (from Section 5.4.2) that efficiency loss in SAMP-SB can be attributable to any of three, not necessarily exclusive, causes: dead ends, failure to form a solution when a positive-valued solution exists, and finding a suboptimal solution. We can infer the percentage of instances exhibiting dead-end suboptimality in SAMP-SB by examining the differences between SAMP-SB-D and SAMP-SB totaled over the Negative, Zero, and Suboptimal columns in Table 1. Decommitment does not affect the contribution of no-solution and suboptimal-solution losses, but helps reveal them by eliminating dead-end suboptimality. Hence, we can infer the percentage of instances exhibiting no-solution and suboptimal-solution suboptimality in SAMP-SB by examining the Zero and Suboptimal columns of SAMP-SB-D, respectively.

Table 2 shows the average efficiency attained by the protocols, factored by network and equilibrium existence (where relevant). We see, from the difference between the SAMP-SB-D and SAMP-SB columns, that dead ends are a significant source of inefficiency. Additionally, existence of competitive equilibrium has a significant effect on the performance of the protocols. In these networks, SAMP-SB-D produces nearly perfect efficiency when competitive equilibrium exists (recall that all studied instances of SIMPLE, BIGGER, and MANY-CONS have equilibria), but is is much less effective when equilibrium does not exist, in fact failing to find any solutions in the no-equilibrium cases of UNBALANCED and GREEDY-BAD.

To check whether these differences in performance are significant, we performed Student's t-Tests for each protocol, comparing the mean efficiencies of instances that admit competitive equilibrium with the means of those instances that do not admit competitive equilibrium. Table 3 shows the results, indicating the p-values that the means of equilibrium and no-equilibrium instances came from the same underlying population. In typical analyses, the null hypothesis that the means are equal is rejected if the p-value is below 0.05. With this threshold, it seems we can safely reject the hypothesis that the mean efficiencies of equilibrium and non-equilibrium instances are the same for

| Network | SAMP-SB | SAMP-SB-D |
|---|---|---|
| SIMPLE | 0.997 | 0.997 |
| UNBALANCED | | |
| – equilibrium exists | 0.867 | 0.990 |
| – no equilibrium exists | −20.080 | 0.000 |
| TWO-CONS, case: | | |
| – equilibrium exists | 0.733 | 0.986 |
| – no equilibrium exists | 0.268 | 0.686 |
| BIGGER | 1.000 | 1.000 |
| MANY-CONS | 0.120 | 0.996 |
| GREEDY-BAD, case: | | |
| – equilibrium exists: | −5.320 | 0.990 |
| – no equilibrium exists: | −18.230 | 0.000 |

Table 2: Average efficiency in each network for the protocols.

| Network | SAMP-SB | SAMP-SB-D |
|---|---|---|
| | | |
| UNBALANCED | $6.27 \times 10^{-30}$ | $8.23 \times 10^{-101}$ |
| TWO-CONS | $5.15 \times 10^{-7}$ | $1.43 \times 10^{-22}$ |
| GREEDY-BAD | $1.41 \times 10^{-1}$ | $8.04 \times 10^{-101}$ |

Table 3: P-values computed with the Student's t-Test. The t-Test compared the means efficiencies of instances that admit competitive equilibrium and those that do not admit competitive equilibrium.

Networks UNBALANCED and GREEDY-BAD. Inspection of the data further supports this conclusion, as SAMP-SB-D essentially always produces zero efficiency, but produces perfect efficiency in many of the instances that do admit competitive equilibrium.

On the face of it, the high SAMP-SB/GREEDY-BAD p-value suggests that we cannot safely reject the hypothesis that the mean efficiencies differ between the equilibrium and no-equilibrium instances of the network. However, inspection of the data indicates that this high probability results from one outlying equilibrium instance with a large negative efficiency. Indeed, the fact that SAMP-SB-D always produces essentially optimal results in instances that admit competitive equilibrium, but predominantly produces suboptimal results in the instances without such equilibria, suggests that it is unlikely that the equilibrium and no-equilibrium means are the same for SAMP-SB in Network GREEDY-BAD.

| Network | % λ-δ-Competitive Equilibrium |
|---|---|
| SIMPLE | 100 |
| UNBALANCED | |
| – equilibrium exists | 88 |
| – no equilibrium exists | 0 |
| TWO-CONS, case: | |
| – equilibrium exists | 83 |
| – no equilibrium exists | 2 |
| BIGGER | 96 |
| MANY-CONS | 17 |
| GREEDY-BAD, case: | |
| – equilibrium exists | 75 |
| – no equilibrium exists | 0 |

Table 4: Percentage of instances in which SAMP-SB attained λ-δ-competitive equilibrium.

Table 4 shows the percentage of instances for which SAMP-SB attained λ-δ-competitive equilibrium in each network. It is straightforward to determine whether λ-δ-competitive equilibrium is attained by observing whether there are any dead ends (Theorem 11). Again, we see a strong connection with the existence of competitive equilibrium. One notable exception is MANY-CONS (which always admits a competitive equilibrium), for which SAMP-SB frequently produced dead ends. We do see that λ-δ-competitive equilibria form in a small percentage of the no-equilibrium TWO-CONS instances, although this is not a prevalent phenomenon with the $\delta_b$ and $\delta_s$ parameters we chose.

Table 5 shows the average efficiency, factored by λ-δ-competitive equilibrium attainment (SAMP-SB and SAMP-SB-D produce the same results when λ-δ-competitive equilibrium is attained). We must be careful in drawing conclusions from these statistics because, for any given network case, there were relatively few or many λ-δ-competitive equilibrium instances (Table 4). Still, we note certain salient trends. The λ-δ-competitive equilibrium runs produce near perfect efficiency, with smaller degrees of inefficiency than specified by the bounds in Theorem 12. Because an allocation produced by SAMP-SB is a λ-δ-competitive equilibrium iff there are no dead ends, we should expect that a significant portion of efficiency loss in non-λ-δ-competitive equilibrium pro-ducing runs of SAMP-SB would be attributable to negative surplus incurred from dead ends. The significant differences between the efficiency of SAMP-SB-D and SAMP-SB shown in the non-$\lambda$-$\delta$-competitive equilibrium column provides evidence for this hypothesis. Indeed, it appears that surplus lost to dead ends (as opposed to suboptimal solution attainment) is the dominant cause of inefficiency when $\lambda$-$\delta$-competitive equilibrium is not attained. In all instances, improvement from decommitment is greater than the difference between the efficiency of SAMP-SB-D and optimal efficiency.

| Network | $\lambda$-$\delta$-Equilibrium Not Found | | $\lambda$-$\delta$-Equilibrium Found |
|---|---|---|---|
| | SAMP-SB | SAMP-SB-D | SAMP-SB/ SAMP-SB-D |
| SIMPLE | N/A | N/A | 0.997 |
| UNBALANCED | | | |
| – equilibrium exists | $-0.248$ | 0.989 | 0.998 |
| – no equilibrium exists | $-20.08$ | 0.000 | N/A |
| TWO-CONS, case: | | | |
| – equilibrium exists | $-0.570$ | 0.920 | 1.000 |
| – no equilibrium exists | 0.130 | 0.707 | 1.000 |
| BIGGER | 0.997 | 1.000 | 1.000 |
| MANY-CONS | $-0.060$ | 0.995 | 1.000 |
| GREEDY-BAD, case: | | | |
| – equilibrium exists | $-24.28$ | 0.960 | 1.000 |
| – no equilibrium exists | $-18.22$ | 0.000 | N/A |

Table 5: Average efficiency in each network for the protocols, factored by $\lambda$-$\delta$-competitive equilibrium attainment.

Table 6 shows the average fraction of available surplus obtained by producers, respectively, in each network. Perhaps surprisingly, in some networks the producers can gain significant surplus with the SAMP-SB-D protocol, even though they are bidding to obtain zero surplus. The reason for this is that a producer's output offer indicates the minimum amount it is willing to accept in exchange for its output. But rising buy offers can cause the price to rise above the producer's output offer. This could happen in cases when it is necessary to block out certain agents to have a feasible allocation in quiescence. Note however, that the decommitment step is needed for the producers to obtain high average surplus. Without decommitment, the average producer surplus can be highly negative, as shown in the SAMP-SB column.

## 7. Related Literature

In this section we discuss literature related to our present work. In Section 7.1 we discuss related literature on price-based analysis and auction theory, and in Section 7.2 we discuss related literature on supply chain formation.

| Network | SAMP-SB | SAMP-SB-D |
|---|---|---|
| SIMPLE | 0.000 | 0.000 |
| UNBALANCED, case: | | |
| – equilibrium exists | −0.041 | 0.082 |
| – no equilibrium exists | −20.09 | 0.000 |
| TWO-CONS, case: | | |
| – equilibrium exists | 0.210 | 0.464 |
| – no equilibrium exists | 0.137 | 0.555 |
| BIGGER | 0.001 | 0.001 |
| MANY-CONS | −0.517 | 0.359 |
| GREEDY-BAD, case: | | |
| – equilibrium exists | −6.08 | 0.137 |
| – no equilibrium exists | −18.11 | 0.000 |

Table 6: Average fraction of available surplus obtained by producers in each network for the protocols.

## 7.1 Price-Based Analysis and Auction Theory

We have shown some special cases for which competitive equilibria exist in task dependency networks (polytree and single-input-producer networks), that SAMP-SB always finds λ-δ-competitive equilibrium in trees, and that a minor variant always finds λ-δ-competitive equilibria in single-input-producer networks. A review of the results in price equilibrium and auction theory reveals that such limited positive results are typical.

It is well-known that given arbitrarily divisible goods and convex utility, cost, and production functions, competitive equilibrium prices exist. If additionally, the gross substitutes condition (which is a generalization of no-complementarities) is met, the classic tâtonnement procedure finds competitive equilibrium in a distributed manner.[6]

When goods are discrete, competitive equilibria exist in exchange (non-production) economies if the gross substitutes conditions are met (Bikhchandani & Mamer, 1997; Gul & Stacchetti, 1999; Kelso & Crawford, 1982). Milgrom (2000) showed that the existence of a single complementarity can be sufficient to preclude equilibrium in exchange economies. Bickhchandani and Mamer (1997) also show existence under a variety of other conditions, which do not appear to have natural interpretations in task dependency networks. In exchange economies, the gross substitutes condition also ensures convergence to (approximately) competitive equilibria with simultaneous ascending auctions (Demange et al., 1986; Gul & Stacchetti, 2000).

That distributed price-based auction protocols can leave agents with undesired goods when their preferences are complementary (e.g., dead ends in a task dependency network), is a widely recognized problem. An alternative approach is to use a ***combinatorial auction***, which mediates negotiation at a single location, performing global matching of combinations of goods based on indivisible bids. This general approach has received much attention in the AI community as of late, motivated in part by techniques for quickly performing the necessary global optimization (Andersson et al., 2000; Leyton-Brown et al., 2000; Sandholm & Suri, 2000).

---

6. The reader may consult a standard microeconomic textbook (Mas-Colell et al., 1995) for details on these results.

Currently, some results on combinatorial equilibria and auctions have been established for one-sided (i.e., buyer only) bidding. Bikhchandani and Ostroy (2002) and Wurman and Wellman (2000), using different combinatorial frameworks, provide positive results on equilibrium existence, and properties thereof. Wurman and Wellman describe a combinatorial auction for their framework. Parkes and Ungar (2000) describe a combinatorial auction that is guaranteed to converge to an efficient allocation if agents follow myopic best-response strategies. By adding an "extend-and-adjust" phase, these authors are able to obtain this allocation in *ex post* Nash equilibrium (Parkes & Ungar, 2002). Ausubel and Milgrom (2002) present a proxy-auction mechanism that obtains efficient allocations with straightforward bidding in equilibrium when goods are substitutes.

In the present work we consider only simple, local, myopic bidding policies. These policies are non-strategic, in that agents do not reason about their effect on the negotiations in an attempt to extract greater surplus. The assumption of non-strategic behavior is plausible when there are a large number of agents. In networks with many agents bidding for individual goods, many parallel branches, or many agents in sequence, the potential contribution of any one agent to the value of a solution is relatively small and there is little to gain by strategic behavior.

Our experiments have shown that, even when producers bid to obtain zero surplus with the specified policy, they can obtain positive surplus in some networks. Nevertheless, in smaller networks, the potential for strategic improvement is more pronounced, and our non-strategic assumption becomes less plausible. A widely studied concept used for analyzing strategic behavior is Bayes-Nash equilibrium.[7] Informally, a set of strategies constitutes a Bayes-Nash equilibrium if no single agent has an incentive to deviate from its strategy, given that all other agents play their Bayes-Nash equilibrium strategies. McAfee and McMillan (1987) and Klemperer (1999) survey the state of knowledge of strategic analysis of auctions in exchange economies. Milgrom (2000) provides insights on some of the fundamental challenges to understanding the agent behavior with complementary preferences. However, definitive results are known only for quite restrictive market structures, and do not encompass two-sided markets with complementarities, never mind the multi-level characteristic of negotiation in task dependency networks. The problem of even specifying the information structure of the extensive form game of simultaneous ascending (M+1)st price auctions in task dependency networks, prerequisite to computing Bayes-Nash equilibria, is well beyond the current state of the art in game-theoretic analysis.

As auction theory currently fails to provide satisfactory guidance for understanding strategic behavior in even moderately complicated domains, some have used tournaments as a framework for developing and evaluating candidate agent strategies. The Santa Fe Double Auction Tournament (Rust et al., 1994) provided some unexpected insights into effective strategies in continuous double auctions, and the recent "TAC" series of trading agent competitions (Wellman et al., 2001b, 2003) encouraged the development of sophisticated agent strategies (Greenwald, 2003; Stone & Greenwald, 2000) for a complex market game.

The Vickrey-Clarke-Groves mechanism (Clarke, 1971; Groves, 1973; Vickrey, 1961), also called the Generalized Vickrey Auction (GVA) (MacKie-Mason & Varian, 1994), is a direct revelation approach, where agents report their valuations on allocations, and the auction computes a lump-sum payment. For the GVA, the solution is the optimal allocation based on the reports, and the payment function is such that it is a dominant strategy for agents to report their true utility. Because of this incentive compatibility and perfect efficiency, the GVA may seem ideal from an economic perspective (although the computation can be intractable). However, the GVA is not budget balanced—when both buyers and sellers bid, the GVA can pay out more money than it takes in. Unfortunately this is unavoidable, as it is impossible to simultaneously ensure efficiency, budget balance, and individual rationality (no agent achieves negative surplus) (Myerson & Satterthwaite, 1983). Recently, Babaioff et al. (Babaioff & Nisan, 2001; Babaioff & Walsh, 2003) described distributed auction mechanisms, based on McAfee's trade reduction auction (1992), that obtain incentive compatibility and budget balance in linear supply chains, at the expense of perfect efficiency. Recent work by Parkes, Kalagnanam, and Eso (2001) explores methods to minimize the deviation from efficiency while maintaining budget balance in two-sided GVA-like mechanisms.

---

7. For a foundational reference, Chapter 7 of Fudenberg and Tirole's game theory text (1998) provides a formal treatment of the strategic issues in auction mechanism design and analysis.

## 7.2 Supply Chain Formation

Supply chain management—the problem of accurately forecasting and planning production and deliveries to meet demand and minimize inventory—is an active field of study in operations research. The problem of management differs from supply chain formation in that the exchange partners in the supply chain are pre-established, and it is assumed that information can be gathered from all agents to effectively optimize global production across the supply chain. In contrast, in this work we approached the problem of automating the process of determining supply chain participants dynamically, under the assumption that information and decision making is decentralized. Readers interested in supply chain management may refer to Kjenstad (1998) for an extensive review.

Relatively less effort has been explicitly devoted to the problem we cast as supply chain *formation*, despite the rhetorical appeals to decentralized and dynamic relation-building commonplace in the popular literature. Nevertheless, as we point out in Section 3, some venerable AI methods—in particular the widely-known CONTRACT NET protocol—can in principle be applied to supply chain formation. As discussed in Section 3, the standard CONTRACT NET does not have mechanisms to resolve nontrivial resource contention, precluding a systematic comparison with SAMP-SB on general network structures. We can, however, compare the protocols on network structures for which a resource contention mechanism is not necessary for CONTRACT NET. It is clear that if agents bid their true costs, then CONTRACT NET with greedy allocation will converge to optimal allocations for trees. The same holds for tree structures relaxed to allow multiple-unit input bids. As we have shown, SAMP-SB is guaranteed to converge to approximately efficient allocations for trees. However, it can be shown that it may not converge to good solutions if multiple unit input bids are allowed. In the latter case, competitive equilibrium may not exist, and and we have observed that equilibrium non-existence can substantially hurt efficiency in SAMP-SB allocations. In contrast, producers may receive different prices for the same good with CONTRACT NET. This discriminatory pricing mechanism makes CONTRACT NET robust in the presence of complementarities, for this class of network structures.

Sandholm (1993) examines a specialization of CONTRACT NET for a generalization of Task Oriented Domains (TODs) (Rosenschein & Zlotkin, 1994). Agents begin with an initial allocation of tasks and negotiate task exchanges until there are no mutually beneficial trades. Sandholm's model allows local constraints on task achievement in that an agent can perform only certain combinations of tasks. However, there is no dependency structure—an agent does not rely on other agents' task achievement in order to accomplish its own tasks. Thus, every locally feasible trade results in a globally feasible allocation, and can be executed immediately and independently of other trades. We cannot generally apply an incremental trading protocol to our task allocation model with subtask dependencies. A local exchange may require reallocation throughout the entire network to maintain production feasibility.

Andersson and Sandholm (1998) find that decommitment protocols increase the quality of the resulting allocations in variants of TODs. With incremental trading, decommitment gives agents the opportunity to engage in other more cost-effective contracts. Andersson and Sandholm also consider decommitment penalties to provide friction in reallocation and to compensate agents whose contracts are broken. We expect that such penalties would be an appropriate extension to the SAMP-SB-D protocol.

Veeramani et al. (Veeramani et al., 1999; Joshi et al., 1999) consider issues arising from simultaneous negotiation of multiple subtasking issues at various levels of a supply chain. In their asynchronous model, agents may have the opportunity to finalize a contract while other negotiations are still pending. This uncertainty induces a complex decision problem for agents that do not wish to overextend their commitments.

Hunsberger and Grosz (2000) study the problem of assigning task performance roles to agents in the SharedPlans collaborative planning framework. The model is based on recipes, which describe the precedence constraints on the execution time across the various sub-tasks that constitute a complex task. Contention for shared resources is not modeled explicitly in the recipe, but individual agents may have additional cost, timing, or other constraints, potentially arising from their individual resource limitations. Hunsberger and Grosz use a combinatorial auction to assign tasks to agents, given the constraints, to produce a high-valued shared plan. They find that limiting task assignment to certain combinations in roles can effect a tradeoff between computational and allocative efficiency.

In other work (Walsh et al., 2000; Walsh, 2001), we have studied strategic behavior of agents bidding in a particular one-shot combinatorial auction within the task dependency network model. We empirically compared the performance of SAMP-SB, SAMP-SB-D, and the combinatorial auction (with strategic bidding). The combinatorial auction eliminates the problem of dead ends by allocating inputs and outputs to producers on an all-or-nothing basis. This advantage notwithstanding, combinatorial auctions may not always be an appropriate mechanism. Since finding any feasible supply chain solution is NP-hard (Walsh et al., 2003), sufficiently large problems will be intractable, even for advanced optimization procedures. Even when the computation is tractable, social factors may limit the authority of any one entity to compute allocations over the entire supply chain.

## 8. Extensions and Future Work

The task dependency network model we propose provides a basis for beginning to understand the automation of supply chain formation. We have discussed some ways to extend the bidding policies in our market protocol to accommodate more general production capabilities and consumer preferences. With these extensions we can model capabilities and preferences on multi-attribute goods (e.g., goods with multiple features such as quality and delivery time, in addition to price and quantity) by simply representing each configuration as a distinct good in the network. However, it is clear that this can explode the number of goods with just a few attributes. To effectively handle greater numbers of attributes would require multiattribute auctions (Bichler, 2001), where multiple inseparable features of an exchange are negotiated simultaneously.

In more realistic scenarios, producers may also have to solve complex internal scheduling, planning, forecasting, or other complex problems in order to evaluate their costs and feasible options. These types of extensions would increase the fidelity of the model, but would have implications for agent bidding policies and computation and convergence speed in market protocols. Despite the best efforts of agents to forecast and plan, agents cannot predict with certainty their operation within a formed supply chain. Sophisticated agents would employ probabilistic reasoning techniques in evaluating their options during negotiation. When unexpected events do occur that impair the operation of a formed supply chain, agents would need protocols for repairing or reforming the supply chain.

We assumed a simple set of non-strategic, myopic bidding policies for the simultaneous ascending auction. Because agents must coordinate input and output bids in a dynamic auction mechanism, understanding strategic bidding behavior is a challenging and unsolved, albeit important problem for future work. It seems likely that significant developments in game-theoretic methodology would be necessary to analytically solve, or even realistically specify, the extensive form games of incomplete information corresponding to asynchronous iterative auctions. In the meantime, to make progress in understanding the performance of auctions, we should consider alternate approaches to developing "good" bidding policies. Tournaments have proven to be effective ways to both encourage smart people to design smart trading policies and to evaluate their relative qualities (Rust et al., 1994; Wellman et al., 2001b). Axelrod (1987) used an evolutionary approach to evaluate populations of strategies, with fixed types, for the iterated prisoners' dilemma. A major challenge in applying an evolutionary approach to the supply chain formation problem is to develop a sufficiently rich, yet reasonably searchable set of agent bidding policies.

We suggested decommitment as a solution to the problem of dead ends in SAMP-SB, and a strategic analysis of the protocol would have to take this phase into account. Because producers could not lose money when decommitment is allowed, we should expect that producers would be more willing to participate, and would also be more aggressive in their bidding. Allowing decommitment begs the question of how to enforce that producers decommit only when they are in dead ends, and also does not address the fact that unilateral decisions for decommitment can potentially break the (possibly desirable) contracts of many other downstream producers. To reduce aggressive bidding and mitigate the potential problems, we could charge penalties to producers who initiate decommitment (Andersson & Sandholm, 1998), perhaps paid to the producers whose output contracts get decommitted. This would reduce spurious decommitments while still allowing an out for producers stuck in costly dead ends.

Finally, we note that the market configuration studied here—separate auctions for each good— represents just one possible partition of the scope of negotiations on the supply chain. At the other extreme, production activity could be mediated by one combinatorial auction mechanism covering the entire supply chain (Walsh et al., 2000). This avoids some coordination pitfalls of the separate auction approach, but imposes other disadvantages associated with imposing a mechanism with global scope. Intermediate configurations, involving multiple auctions for clusters of highly related goods, represent a promising alternative for further investigation.

ACKNOWLEDGMENTS

This paper includes material previously presented at the Sixteenth International Joint Conference on Artificial Intelligence (IJCAI-99) (Walsh & Wellman, 1999). This work was supported in part by NSF grant IIS-0205435.

## Appendix A. Proofs

This appendix provides proofs of our theorems. For convenience, we restate the theorems before the proofs.

In our proofs, it is sometimes useful to index the position of a producer in a network. The **C-level of a producer** with output $g$ is the maximum distance from it to any consumer, formally stated as follows: one if no producer, but some consumer, has $g$ as input, and $k + 1$ if the maximum level of any producer with input $g$ is $k$. The **S-level of a producer** is defined similarly, but with respect to the distance to any producer with no input, and with a basis of zero for producers that have no inputs themselves. Both the C-level and S-level are well defined, by acyclicity.

### A.1 Proof of Theorem 1

Let $(V, E)$ be a network with no input complementarities, that is all producers have at most one input, and let $(V^*, E^*)$ be the optimal allocation for $(V, E)$. For convenience, we partition producers $\Pi$ into sets $\Pi_1$, the producers with a single input, and $\Pi_0$, the producers with no inputs.

Procedure `No Input Complementarities Equilibrium` constructs prices that support a competitive equilibrium for $(V^*, E^*)$.

`No Input Complementarities Equilibrium:`

1. Initialize all prices to zero.

2. Perform any of the following until no price changes are made:

   (a) If for some $c \in C \setminus V^*$, we have $v_c(g) > p(g)$, where $\langle g, c \rangle \in E \setminus E^*$,
       $p(g) \leftarrow v_c(g)$.

   (b) If for some $c \in C \cap V^*$, we have $v_c(g') - p(g') > v_c(g) - p(g) \geq 0$,
       where $\langle g, c \rangle \in E^*$ and $\langle g', c \rangle \in E \setminus E^*$,
       $p(g') \leftarrow v_c(g') - (v_c(g) - p(g))$.

   (c) If for some $\pi \in \Pi_0 \cap V^*$, we have $p(g_\pi) < \kappa_\pi$, where $\langle \pi, g_\pi \rangle \in E^*$,
       $p(g_\pi) \leftarrow \kappa_\pi$.

   (d) If for some $\pi \in \Pi_1 \cap V^*$, we have $p(g_\pi) < p(g) + \kappa_\pi$
       where $\langle \pi, g_\pi \rangle \in E^*$ and $\langle g, \pi \rangle \in E^*$,
       $p(g_\pi) \leftarrow p(g) + \kappa_\pi$.

   (e) If for some $\pi \in \Pi_1 \setminus V^*$, we have $p(g_\pi) > p(g) + \kappa_\pi$,
       where $\langle \pi, g_\pi \rangle \in E \setminus E^*$ and $\langle g, \pi \rangle \in E \setminus E^*$,
       $p(g) \leftarrow p(g_\pi) - \kappa_\pi$.

In network $(V, E)$ (with no input complementarities) at prices $p$, a **closed, reverse-surplus sequence** is a directly connected sequence of agents and goods such that every agent would be better off by "reversing" its allocation. Formally, it is a sequence $(n_1, \ldots, n_k)$ of vertices $n_i \in V$, such that:

1. $\langle n_i, n_{i+1} \rangle \in E$ or $\langle n_{i+1}, n_i \rangle \in E$ for all $i$ where $1 \leq i \leq k-1$.

2. $n_k \in G$.

3. $n_1 \in (C \setminus V^*) \cup (\Pi_0 \cap V^*)$.

    (a) If $n_1 \in C \setminus V^*$, then $\langle n_2, n_1 \rangle \in E \setminus E^*$ and $n_1$ would obtain nonnegative surplus at $p$ from obtaining $n_2$. If $1 = k-1$, then $n_1$ would obtain strictly positive surplus at $p$ from obtaining $n_2$.

    (b) If $n_1 \in \Pi_0 \cap V^*$, then $\langle n_1, n_2 \rangle \in E^*$ and $n_1$ would obtain nonpositive surplus at $p$ from being active. If $k = 2$, then $n_1$ would obtain strictly negative surplus at $p$ from being active.

4. For $i \geq 2$, if $n_i \in A$ then $n_i \in \Pi_1 \cup (C \cap V^*)$.

    (a) If $n_i \in \Pi_1 \cap V^*$, then $\langle n_{i-1}, n_i \rangle \in E^*$, $\langle n_i, n_{i+1} \rangle \in E^*$, and $n_i$ would obtain nonpositive surplus at $p$ from being active. If $i = k-1$, then $n_i$ would obtain strictly negative surplus at $p$ by being active.

    (b) If $n_i \in \Pi_1 \setminus V^*$, then $\langle n_{i+1}, n_i \rangle \in E^*$, $\langle n_i, n_{i-1} \rangle \in E^*$, and $n_i$ would obtain nonnegative surplus at $p$ from being active. If $i = k-1$, then $n_i$ would obtain strictly positive surplus at $p$ by being active.

    (c) If $n_i \in C \cap V^*$, then $\langle n_{i-1}, n_i \rangle \in E^*$, $\langle n_{i+1}, n \rangle \in E \setminus E^*$, and $n_i$ would obtain no less surplus from $n_{i+1}$ than from $n_{i-1}$ at $p$. If $i = k-1$, then $n_i$ would obtain strictly better surplus from $n_{i+1}$ than from $n_{i-1}$.

An **open reverse-surplus** sequence is the same as a closed, reverse-surplus sequence except that, instead of Condition 3, we have $n_1 \in G$ and $n_2 \in \Pi_1 \cup (C \cap V^*)$ as with Condition 4. Clearly any closed, reverse-surplus sequence of length greater than two contains an open, reverse-surplus sequence.

**Lemma 18** *Procedure* `No Input Complementarities Equilibrium` *does not reach a state such that there is an open, reverse-surplus sequence $K = (n_1, \ldots, n_k)$ constituting a cycle with $n_1 = n_k$ and $k \geq 3$.*

*Proof.* Assume, contrary to which we wish to prove, that there is such a cycle $K$ at prices $p$. Moreover, let the cycle be the smallest, in that it contains no other such cycle.

We show how to create an alternate, feasible solution $(V', E')$ with a higher value than $(V^*, E^*)$, giving us a contradiction. Initialize $(V', E') = (V^*, E^*)$. For all $n_j$, where $1 \leq j < k$, if $\langle n_j, n_{j+1} \rangle \in E^*$, remove the edge from $E'$, and if the edge is in $E \setminus E^*$, add the edge to $E'$. Also, add and remove vertices as necessary to be consistent with the added and removed edges.

Each producer in $(V', E')$ is feasible because it was feasible in $(V^*, E^*)$ and if it has an input, either both its input and output are added, both are removed, or neither is changed. Consider a good $n_j \in G$, with $1 < j < k$. Since $j - 1 > 1$, it must be that agents $n_{j-1}$ and $n_{j+1}$ are in $\Pi_1 \cup (C \cap V^*)$. By inspecting Conditions 4(a)–4(b) of the definition of a closed, reverse-surplus sequence (which also apply to an open reverse-surplus sequence), we see that edges incident on $n_j$ are added or removed in such a way that $n_j$ is in material balance. Similarly, considering agents $n_{k-1}$, $n_2$, and good $n_1 = n_k$, we have material balance for good $n_1 = n_k$. Since goods are in material balance and producers are feasible, $(V', E')$ is feasible.

The surpluses of agents not in $K$ are unaffected by the transformation. By definition of an open, reverse-surplus sequence, every agent in $K$ obtains no lower surplus at $p$ after the transformation, and agent $n_{k-1}$ obtains strictly higher surplus at $p$. Because the value of a feasible allocation is the sum of agent surpluses at any particular prices (Lemma 22), we must then have value$((V', E')) >$ value$((V^*, E^*))$. But this contradicts the optimality of $(V^*, E^*)$, so the assumption that $K$ exists must be false. $\square$

**Lemma 19** *If the price of good $n_k$ increases in Procedure* No Input Complementarities Equilibrium, *then there there exists a finite closed, reverse-surplus sequence $(n_1, \ldots, n_k)$ at prices $p$ just before the price increase.*

*Proof.*

We show how to construct the desired closed, reverse-surplus sequence, referring to the conditions in the definition, and to the steps in Procedure No Input Complementarities Equilibrium. The price increase of $n_k$ occurred in one of Steps 2(a)–2(e), triggered by agent $n_{k-1}$. Since the step was triggered, $n_{k-1}$ would obtain strictly better surplus by reversing its allocation at $p$, as specified in the conditions 3(a), 3(b), or 4(a)–4(c). If the price of $n_k$ was increased in Step 2(a) or 2(c), then we have the desired closed, reverse-surplus sequence, with $n_{k-1} \in (C \setminus V^*) \cup (\Pi_0 \cap V^*)$ and $k - 1 = 1$. Otherwise, the price or $n_k$ was increased in Step 2(b), 2(d), or 2(e), with $n_{k-1} \in \Pi_1 \cup (C \cap V^*)$ and $k - 1 > 2$. In this case, we let $n_{k-2}$ be the good that also matched the condition of the step.

If the price of $n_{k-2}$ increased, then Procedure No Input Complementarities Equilibrium ensures that we can find an agent $n_{k-3}$ matching one of the Conditions 3(a), 3(b), or 4(a)–4(c). If, on the other hand, $p(n_{k-2}) = 0$, then because producers have positive costs and consumers have positive values, we can also find such an agent $n_{k-3}$. If we find an agent that corresponds to condition 3(a) or 3(b), then $k - 3 = 1$ and we are done. Otherwise, we can find a good $n_{k-4}$, as we did $n_{k-2}$, and continue in the same manner.

Clearly, this process constructs an open, reverse-surplus sequence. Now, we must show that this process of selecting vertices eventually selects an element $n_1 \in (C \setminus V^*) \cup (\Pi_0 \cap V^*)$. Since $(V, E)$ is finite, and since by Lemma 18 there can be no cycles in any open, reverse-surplus sequence, we must eventually find a $n_1 \in (C \setminus V^*) \cup (\Pi_0 \cap V^*)$ to give us a closed, reverse-surplus sequence. $\square$

**Lemma 20** *Procedure* No Input Complementarities Equilibrium *terminates.*

*Proof.* Assume, contrary to which we wish to prove, that the procedure does not terminate and that the price of good $g$ increases an infinite number of times. Consider a cycle $K = (n_1 = g, \ldots, n_k = g)$ of vertices $n_i \in V$, $k \geq 3$ such that:

1. $\langle n_i, n_{i+1} \rangle \in E$ or $\langle n_{i+1}, n_i \rangle \in E$ for all $i \in \{1, \ldots, k-1\}$.

2. For all $i \in \{2, \ldots, k-1\}$, $n_i \neq g$.

3. For all $i \in \{3, \ldots, k\}$, if $n_i \in G$, the price increase of good $n_i$ occurred in one of the Steps 2(b), 2(d), or 2(e) in the procedure, and agent $n_{i-1}$ and good $n_{i-2}$ also matched the condition in that step. Furthermore, the price increase of $n_{i-2}$, triggered by agent $n_{i-3}$ and good $n_{i-4}$, caused the need for the price increase of good $n_i$.

Because the price of $g$ increases an infinite number of times, such a cycle must exist.

Let $p$ be prices such that $p(n_1)$ is as it was when $n_1$ and agent $n_2$ triggered the price increase of $n_3$, and for all $n_i \in G$ and $1 < i < k$, $p(n_i)$ is as it was just after the it was increased, as triggered by agent $n_{i-1}$ and good $n_{i-2}$. The price of all other goods is an arbitrary nonnegative number.

By the way we constructed $p$, and by the way prices are increased in the procedure, $K$ must be an open, reverse-surplus sequence. But by Lemma 18, such a $K$ cannot exist. Therefore, the procedure terminates. $\square$

**Theorem 1** *Competitive equilibria exist for any network with no input complementarities.*

*Proof.* We show that Procedure `No Input Complementarities Equilibrium` terminates at prices $p$ with every agent obtaining its maximum surplus according to $(V^*, E^*)$. Since $(V^*, E^*)$ is efficient, it is also feasible, giving us a competitive equilibrium at prices $p$.

By Lemma 20, the procedure terminates. Clearly, when the procedure terminates, agents in $\Pi_1 \cup (\Pi_0 \cap V^*) \cup (C \setminus V^*)$ optimize according to $(V^*, E^*)$. It remains to show the same for $(C \cap V^*) \cup (\Pi_0 \setminus V^*)$. Assume, contrary to which we wish to prove, that some $a \in (C \cap V^*) \cup (\Pi_0 \setminus V^*)$ does not optimize according to $(V^*, E^*)$.

Consider the case where $a \in (C \cap V^*)$ and $\langle g, a \rangle \in E^*$. Since the algorithm guarantees that $a$ does not prefer any other good $g'$ to $g$ at prices $p$, it must be that $p(g) > v_c(g)$. Let $p'$ be the prices immediately before the price of $g$ rose above $v_c(g)$ and $p''$ be the prices immediately after. By Lemma 19, there is a closed, reverse-surplus sequence $(n_1, \ldots, n_k = g)$ at prices $p'$. At $p''$, the conditions of the closed, reverse-surplus sequence hold, except that the surplus condition in 4(a), 4(b), or 4(c) that applies to $n_{k-1}$ becomes non-strict. However, $a$ obtains a strictly negative surplus at $p''$. Denote $a$ as $n_{k+1}$.

We can create an alternate, feasible solution $(V', E')$ as in the proof of Lemma 18 by adding edges $\langle n_i, n_{i+1} \rangle$ that are in $E \setminus E^*$, and removing such edges that are in $E^*$, for all $i \in \{1, \ldots, k\}$. The surpluses of agents not in $K$ are unaffected by the transformation. Every agent in $(a_1, \ldots, n_{k-1})$ obtains no lower surplus at $p''$ after the transformation. Agent $a = n_k$ obtains zero surplus after the transformation, which is higher than the negative surplus it had before. Because the value of a feasible allocation is the sum of agent surpluses at any particular prices (Lemma 22), we must have value$((V', E')) >$ value$((V^*, E^*))$. But this contradicts the optimality of $(V^*, E^*)$, so it must be that $p(g) \le v_c(g)$ and $a$ is obtaining its maximum surplus at $p$ in $(V^*, E^*)$.

If, on the other hand, $a \in (\Pi_0 \setminus V^*)$, and $\langle a, g \rangle \in E$. It must be that $\kappa_a < p(g)$. We can use the same line of proof as the case of $C \cap V^*$ to show that $(V^*, E^*)$ has a suboptimal value, providing a contradiction. Thus $a$ must optimize according to $(V^*, E^*)$ at $p$.

Thus we have shown that the algorithm terminates with all agents optimizing according to $(V^*, E^*)$ at $p$. Thus $p$ supports a competitive equilibrium for allocation $(V^*, E^*)$. $\square$

## A.2 Proof of Theorem 2

Given a polytree $(V, E)$ and an efficient allocation $(V^*, E^*)$, we present Procedure `Polytree Equilibrium` that constructs lower bounds $p_-(g)$ and upper bounds $p^-(g)$ on the prices of all goods $g$, and in turn uses these bounds to construct prices $p$ for all goods. Then we prove that the resulting prices are in fact competitive equilibrium prices that support $(V^*, E^*)$.

Observe that, for the purposes of competitive equilibrium pricing, we can treat a consumer $c$ that wishes to obtain one good from the set $G_c$ as a consumer that desires a single good $g_c$ with value $v_c(g_c) = v_c = \max_{g \in G_c} v_c(g)$, along with additional producers. For each $g \in G_c$ we create a producer $\pi$ with output $g_c$, input $g$, and with $\kappa_\pi = v_c - v_c(g)$. Thus, without loss of generality, we consider only consumers with preferences for single goods. We denote as $g_c$ the good that consumer $c$ desires and denote as $v_c$ the value $c$ has for $g_c$.

We refer to all $n' \in V$ such that either $\langle n, n' \rangle \in E$ or $\langle n', n \rangle \in E$ as the ***neighbors*** of a vertex $n \in V$. We use $\perp$ to refer to a null vertex that is not a neighbor of any other vertex.

`Polytree Equilibrium`:

1. For each $g \in G$, $p_-(g) \leftarrow 0$ and $p^-(g) \leftarrow \infty$.

2. For each connected subgraph $(\hat{V}, \hat{E}) \subseteq (V, E)$, select a $g \in G \cap \hat{V}$ arbitrarily:
   Perform `Set Bounds(`$g$`, ` $\perp$`)`.
   $p(g) \leftarrow p_-(g)$.

`Set Bounds` recursively visits the vertices, updating the price bounds in postorder (i.e., as the recursion unwinds) and setting prices to either the lower or upper bounds. Because $(V, E)$ is a polytree, the procedure sets the price for each good exactly once.

In `Set Bounds(`$n$`, ` $r$`)`, if $n \in A$, then $r \in G$ and the procedure either updates $p_-(r)$ or $p^-(r)$ after the bounds for all neighbors of $n$, other than $r$, have been fixed. If it updates $p_-(r)$, it does so in such a way that if $n \notin V^*$ then $n$, if active, would get a nonpositive surplus for any $p(r) \geq p_-(r)$, given the bounds on the other neighbors of $n$, and if $n \in V^*$ then $n$, if active, would get a nonnegative surplus for any $p(r) \geq p_-(r)$, given the bounds on the other neighbors of $n$. Since $p_-(r)$ only increases (Steps 2, 4(b), and 5(c)), this property is maintained. Similarly, if `Set Bounds(`$n$`, ` $r$`)` updates $p^-(r)$, it does so in such a way that if $n \notin V^*$ then $n$, if active, would get a nonpositive surplus for any $p(r) \leq p^-(r)$, given the bounds on the other neighbors of $n$, and if $n \in V^*$ then $n$, if active, would get a nonnegative surplus for any $p(r) \leq p^-(r)$, given the bounds on the other neighbors of $n$. Since $p^-(r)$ only decreases (Steps 3, 4(c), and 5(b)), this property is maintained.

`Set Bounds(`$n$`, ` $r$`)`:

1. For each neighbor $z$ of $n$ such that $z \neq r$, perform `Set Bounds(`$z$`, ` $n$`)`.

2. If $n \in C \setminus V^*$,
   $p_-(r) \leftarrow \max(v_n, p_-(r))$.

3. Else if $n \in C \cap V^*$,
   $p^-(r) \leftarrow \min(v_n, p^-(r))$.

4. Else if $n \in \Pi \setminus V^*$ then,

   (a) For each neighbor $g$ of $n$ such that $g \neq r$
       If $g$ is an input of $n$
       $p(g) \leftarrow p^-(g)$.
       Else $g$ is the output of $n$,
       $p(g) \leftarrow p_-(g)$.

(b) If $r$ is an input of $n$,
    and with the output, $g_n$, of $n$
$$p_-(r) \leftarrow \max(p_-(r),\ p_-(g_n) - \textstyle\sum_{\langle g,n \rangle \in E,\ g \neq r} p^-(g) - \kappa_n).$$

(c) Else $r$ is the output of $n$,
$$p^-(r) \leftarrow \min(p^-(r),\ \textstyle\sum_{\langle g,n \rangle \in E} p^-(g) + \kappa_n).$$

5. Else if $n \in \Pi \cap V^*$ then,

    (a) For each neighbor $g$ of $n$ such that $g \neq r$,
        If $g$ is an input of $n$,
$$p(g) \leftarrow p_-(g).$$
        If $g$ is the output of $n$,
$$p(g) \leftarrow p^-(g).$$

    (b) If $r$ is an input of $n$,
        and with the output, $g_n$, of $n$),
$$p^-(r) \leftarrow \min(p^-(r),\ p^-(g_n) - \textstyle\sum_{\langle g,n \rangle \in E,\ g \neq r} p_-(g) - \kappa_n).$$

    (c) Else $r$ is the output of $n$,
$$p_-(r) \leftarrow \max(p_-(r),\ \textstyle\sum_{\langle g,n \rangle \in E} p_-(g) + \kappa_n)$$

**Lemma 21** *Procedure* `Polytree Equilibrium` *computes price bounds such that* $p_-(g) \leq p^-(g)$ *for all goods* $g \in G$.

*Proof.* Assume, contrary to which we wish to prove, that at some state there is some $\hat{g} \in G$ such that $p_-(\hat{g}) > p^-(\hat{g})$. Assume further that $\hat{g}$ is the first such good visited.

We say that agent $a$ constrained $p_-(g)$ if `Set Bounds`$(a,\ g)$ was the last to change $p_-(g)$. Similarly, we say that agent $a$ constrained $p^-(g)$ if `Set Bounds`$(a,\ g)$ was the last to change $p^-(g)$.

Recall from Lemma 22 that the value of any feasible allocation is equal to the sum of the agent surpluses at any particular prices. We show how to transform $(V^*, E^*)$ to an alternate feasible allocation $(V', E')$ and compute alternate prices $p$ to show that the sum of surpluses in $(V', E')$ is greater than in $(V^*, E^*)$.

First, initialize $(V', E') = (V^*, E^*)$ and for each good $g \in G$ initialize $p(g) = 0$. Next, set $p(\hat{g}) = p_-(\hat{g})$. Then we recursively change prices and the allocation for a portion of the subtree rooted at $\hat{g}$. Perform `Lower Bound`$(a,\ \hat{g})$ for the agent $a$ that constrained $p_-(\hat{g})$ and perform `Upper Bound`$(a,\ \hat{g})$ for the agent that constrained $p^-(g)$.

Throughout the transformation, we perform `Lower Bound`$(\tilde{a},\ \tilde{g})$ iff we visit $\tilde{g}$ and agent $\tilde{a}$ constrained $p_-(\tilde{g})$. Similarly, we perform `Upper Bound`$(\tilde{a},\ \tilde{g})$ iff we visit $\tilde{g}$ and agent $\tilde{a}$ constrained $p^-(\tilde{g})$. The following describes these portions of the transformation.

`Lower Bound`$(a,\ g)$:

1. If $a \in \Pi \setminus V^*$, it must be that $g$ is an input of $a$ (because $a$ constrained $p_-(g)$).
    For each neighbor $g' \neq g$ of $a$:

   (a) If $g'$ is an input of $a$,
       $$p(g') \leftarrow p^-(g'),$$
       perform `Upper Bound`$(a',\ g')$ for the agent $a'$ that constrained $p^-(g')$.

   (b) Else ($g'$ is an output of $a$),
       $$p(g') \leftarrow p_-(g'),$$
       perform `Lower Bound`$(a',\ g')$ for the agent $a'$ that constrained $p_-(g')$.

2. Else if $a \in \Pi \cap V^*$, it must be that $g$ is an output of $a$ (because $a$ constrained $p_-(g)$).
   For each input $g'$ of $a$:
       $$p(g') \leftarrow p_-(g'),$$
       perform `Lower Bound`$(a',\ g')$ for the agent $a'$ that constrained $p_-(g')$.

3. If $a \in V^*$,
       remove $a$ and all incident edges from $(V', E')$.

4. Else if $a \in V \setminus V^*$,
       add $a$ and all incident edges to $(V', E')$.

`Upper Bound`$(a,\ g)$:

1. If $a \in \Pi \setminus V^*$, it must be that $g$ is an output of $a$ (because $a$ constrained $p^-(g)$).
   For each input $g'$ of $a$:
       $$p(g') \leftarrow p^-(g'),$$
       perform `Upper Bound`$(a',\ g')$ on the agent $a'$ that constrained $p^-(g')$.

2. If $a \in \Pi \cap V^*$, it must be that $g$ is an input of $a$ (because $a$ constrained $p^-(g)$).
   For each neighbor $g' \neq g$ of $a$:

   (a) If $g'$ is an input of $a$,
       $$p(g') \leftarrow p_-(g'),$$
       perform `Lower Bound`$(a',\ g')$ for the agent $a'$ that constrained $p_-(g')$.

   (b) Else ($g'$ is an output of $a$),
       $$p(g') \leftarrow p^-(g'),$$
       perform `Upper Bound`$(a',\ g')$ for the agent $a'$ that constrained $p^-(g')$.

3. If $a \in V^*$,
       remove $a$ and all incident edges from $(V', E')$.

4. Else if $a \in V \setminus V^*$,
       add $a$ and all incident edges to $(V', E')$.

   Observe that, because $(V, E)$ is a polytree, a vertex can be visited at most once by either `Upper Bound` or `Lower Bound`.

   Now we show that $(V', E')$ is feasible. Consumers are always feasible. Producers are feasible because we add or remove all incident edges when we add or remove a producer, respectively. We now prove that every $g \in G$ is in material balance in $(V', E')$.

Consider good $\hat{g}$ for which $p_-(\hat{g}) > p^-(\hat{g})$. `Lower Bound`$(a, \hat{g})$ is performed only if agent $a$ constrained $p_-(\hat{g})$, which occurred in 2, 4(b), or 5(c) of `Set Bounds`$(a, \hat{g})$. Therefore `Lower Bound`$(a, \hat{g})$ either adds $\langle \hat{g}, a \rangle \in E \setminus E^*$ or else removes $\langle a, \hat{g} \rangle \in E^*$. `Upper Bound`$(a, \hat{g})$ is performed only if $a$ constrained $p^-(\hat{g})$, which occurred in 3, 4(c), or 5(b) of `Set Bounds`$(a, \hat{g})$. Therefore `Upper Bound`$(a, g)$ either adds $\langle a, \hat{g} \rangle \in E \setminus E^*$ or else removes $\langle \hat{g}, a \rangle \in E^*$. For any possible combination, material balance is maintained for $\hat{g}$.

Now consider any other good $g \neq \hat{g}$. If $g$ is visited by `Lower Bound`$(a, g)$, then $p(g)$ was set to $p_-(g)$ in one of the following ways, immediately prior:

1. $p(g)$ was set to $p_-(g)$ by 1(b) of `Lower Bound`$(\tilde{a}, \tilde{g})$, for some other agent $\tilde{a} \in \Pi \setminus V^*$ and other good $\tilde{g}$. In this case, $g$ is an output of $\tilde{a}$ so $\langle \tilde{a}, g \rangle \in E \setminus E^*$ was added to $(V', E')$ in 4 of `Lower Bound`$(\tilde{a}, \tilde{g})$.

2. $p(g)$ was set to $p_-(g)$ by 2 of `Lower Bound`$(\tilde{a}, \tilde{g})$, for some other agent $\tilde{a} \in \Pi \cap V^*$ and other good $\tilde{g}$. In this case $g$ is an input of $\tilde{a}$ so $\langle g, \tilde{a} \rangle \in E^*$ was removed from $(V', E')$ in 3 of `Lower Bound`$(\tilde{a}, \tilde{g})$.

3. $p(g)$ was set to $p_-(g)$ by 2(a) of `Upper Bound`$(\tilde{a}, \tilde{g})$, for some other agent $\tilde{a} \in \Pi \cap V^*$ and other good $\tilde{g}$. In this case case $g$ is an input of $\tilde{a}$ so $\langle g, \tilde{a} \rangle \in E^*$ was removed from $(V', E')$ in 3 of `Upper Bound`$(\tilde{a}, \tilde{g})$.

One of the following operations occurred in `Lower Bound`$(a, g)$:

1. If $a$ constrained $p_-(g)$ in 2 or 4(b) of `Set Bounds`$(a, g)$, then $\langle g, a \rangle \in E \setminus E^*$ is added to $(V', E')$ in 4 of `Lower Bound`$(a, g)$.

2. If agent $a$ constrained $p_-(g)$ in 5(c) of `Set Bounds`$(a, g)$, then $\langle a, g \rangle \in E^*$ is removed from $(V', E')$ in 3 of `Lower Bound`$(a, g)$.

For any possible combination of additions or removals of edges incident on $g$ prior to, and in `Lower Bound`$(a, g)$, material balance is maintained for $g$. We can show a similar result if $g$ is visited by `Upper Bound`$(a, g)$. Hence we have established feasibility of $(V', E')$.

Now we show that for any agent $a \in A$, $\sigma(a, (V', E'), p) \geq \sigma(a, (V^*, E^*), p)$, and there is some agent $a' \in A$ such that $\sigma(a', (V', E'), p) > \sigma(a', (V^*, E^*), p)$.

For any agent $a$ not visited in the construction of $(V', E')$, $\sigma(a, (V', E'), p) = \sigma(a, (V^*, E^*), p)$, because $a$ has the same allocation as in $(V^*, E^*)$.

Consider $a'$ visited by `Upper Bound`$(a', \hat{g})$. Because $a'$ was thus visited, $a'$ constrained $p^-(\hat{g})$. `Upper Bound`$(a', \hat{g})$ sets the prices of all other neighbor goods $g \neq \hat{g}$ to the prices used to compute $p^-(\hat{g})$ in `Set Bounds`$(a', \hat{g})$. The prices of these neighboring goods were computed such that if $a' \in V^*$, $a'$ would get negative surplus at any price for $\hat{g}$ above $p^-(\hat{g})$ and if $a' \in V \setminus V^*$ it would get a positive surplus for at any price for $\hat{g}$ above $p^-(\hat{g})$. But, in the alternate prices we computed, $p(\hat{g}) = p_-(\hat{g})$, and we assume $p_-(\hat{g}) > p^-(\hat{g})$. Since $a'$ is in $V'$ if and only if it is not in $V^*$, we have $\sigma(a', (V', E'), p) > \sigma(a', (V^*, E^*), p)$.

Now consider any other $a \in A$, $a \neq a'$, visited in the construction of $(V', E')$. If $a$ is visited by `Upper Bound`$(a, g)$, then $p(g) = p^-(g)$ and $a$ must have constrained $p^-(g)$. If $a \in C$, then `Set Bounds`$(a, g)$ set $p^-(g)$ such that $v_a - p^-(g) = 0$. If $a \in \Pi$, `Upper Bound`$(a, g)$ sets the prices of the other goods neighboring $a$ to the prices used to compute $p^-(g)$ in `Set Bounds`$(a, g)$. These neighboring prices are such that if $a$ were active and feasible, it would get zero surplus when $p(g) = p^-(p)$. Thus $\sigma(a, (V', E'), p) = \sigma(a, (V^*, E^*), p)$. If, on the other hand, $a$ is visited by `Lower Bound`, then $p(g) = p_-(g)$ and $a$ must have constrained $p_-(g)$. By a similar argument we used with `Upper Bound`, $a$ gets a zero surplus when $p(g) = p_-(g)$. Again, this gives $\sigma(a, (V', E'), p) = \sigma(a, (V^*, E^*), p)$.

We have shown that for any agent $a \in A$, $\sigma(a, (V', E'), p) \geq \sigma(a, (V^*, E^*), p)$, and there is some agent $a' \in A$ such that $\sigma(a', (V', E'), p) > \sigma(a', (V^*, E^*), p)$. But then value$((V', E')) >$ value$((V^*, E^*))$, which is a contradiction. Hence, the initial assumption that $p_-(\hat{g}) > p^-(\hat{g})$ must be false. But then $p_-(g) \leq p^-(g)$ for all goods $g$.

□

**Theorem 2** *Competitive equilibria exist for any polytree.*

*Proof.* We show that agents optimize according to $(V^*, E^*)$ at the prices $p$ computed by procedure `Polytree Equilibrium`. Since $(V^*, E^*)$ is feasible by definition, the resulting prices and allocation constitute a competitive equilibrium for $(V, E)$.

Because the construction of $p_-(g)$ ensures that it never decreases, Step 2 of `Set Bounds` ensures that every consumer $c \in C \setminus V^*$ optimizes if $p(g_c) \geq p_-(g_c)$. Because $p_-(g_c) \leq p(c) \leq p^-(g_c)$ (by construction of $p$ and by Lemma 21), $c$ then optimizes according to $(V^*, E^*)$. By a similar argument, every $c \in C \cap V^*$ optimizes according to $(V^*, E^*)$.

Consider a producer $\pi \in \Pi \setminus V^*$, visited by `Set Bounds(`$\pi$`, `$g$`)`. If good $g$ is an input of $\pi$, then 4(a) of `Set Bounds(`$\pi$`, `$g$`)` sets the price of every other neighbor good $g' \neq g$ of $\pi$ to the price bounds used to compute $p_-(g)$ in Step 4(b) of `Set Bounds(`$\pi$`, `$g$`)`. Moreover, $p_-(g)$ is set to the smallest price such that $\pi$ could get a maximum surplus of zero, given the specified bounds of the other neighbor goods. Since $p_-(g)$ could only increase subsequently, since $p_-(g) \leq p(g) \leq p(g)$ (by the construction of $p$ and by Lemma 21), and since the price of each good is set only once (because $(V, E)$ is a polytree) $\pi$ cannot get a positive surplus at the prices set by `Set Bounds(`$\pi$`, `$g$`)`. By a similar argument, if $g$ is an output of $\pi$, then in Step 4(c) $p^-(g)$ is set to the largest price such that $\pi$ would get at a maximum surplus of zero, given the prices set on the neighbor goods. Since again, $p_-(g) \leq p(g) \leq p^-(g)$, $p^-(g)$ only increases subsequently, and the price of each good is set only once, it must be that $\pi$ can get at most zero surplus. Thus $\pi$ optimizes according to $(V^*, E^*)$. Symmetrically, we can see that every $\pi \in \Pi \cap V^*$ optimizes according to $(V^*, E^*)$.

We have shown that all agents optimize according to $(V^*, E^*)$ at $p$, hence we have shown that a competitive equilibrium exists for polytree $(V, E)$. □

## A.3 Proof of Theorem 3

**Lemma 22** *The value of a feasible allocation $(V', E')$, at any prices $p$, can be expressed as:*

$$value((V', E')) = \sum_{a \in A} \sigma(a, (V', E'), p). \tag{1}$$

*Proof.* Equation (1) expands to:

$$\text{value}((V', E')) = \sum_{c \in C} \left( v_c((V', E')) - \sum_{\langle g, c \rangle \in E'} p(g) \right)$$
$$+ \sum_{\pi \in \Pi} \left( \sum_{\langle \pi, g \rangle \in E'} p(g) - \sum_{\langle g, \pi \rangle \in E'} p(g) - \kappa_\pi((V', E')) \right).$$

Since all goods are in material balance in a feasible allocation, all the price terms cancel out. We are left with

$$\sum_{c \in C} v_c((V', E')) - \sum_{\pi \in \Pi} \kappa_s((V', E')),$$

which is the original formula for the value of a solution (Definition 1). $\square$

**Theorem 3** *If $(V', E')$ is a $\lambda$-$\delta$-competitive equilibrium for $(V, E)$ at some prices $p$, then $(V', E')$ is a feasible allocation with a nonnegative value that differs from the value of an efficient allocation by at most $\sum_{\pi \in \Pi} [\sum_{\langle g, \pi \rangle \in E} \lambda_\pi^g + \delta_s] + |C| \delta_b$.*

*Proof.* We refer to the four conditions for a $\lambda$-$\delta$-competitive equilibrium (Definition 4). Let $(V^*, E^*)$ be an efficient allocation for $(V, E)$.

Conditions (3) and (4) imply that $(V', E')$ is feasible. Recall the formula for the value of a feasible allocation from Equation (1). Since $(V', E')$ and $(V^*, E^*)$ are both feasible, we can express their values as

$$\text{value}((V', E')) \quad = \quad \sum_{a \in A} \sigma(a, (V', E'), p), \tag{2}$$

$$\text{value}((V^*, E^*)) \quad = \quad \sum_{a \in A} \sigma(a, (V^*, E^*), p). \tag{3}$$

For all $c \in C$, by Condition (2), $\sigma(c, (V', E'), p) \geq H_c(p) - \delta_b$. Because no allocation is any better for an agent than its optimal allocation, $\sigma(c, (V^*, E^*), p) \leq H_c(p)$. Thus,

$$\sigma(c, (V^*, E^*), p) - \sigma(c, (V', E'), p) \quad \leq \quad \delta_b. \tag{4}$$

For all $\pi \in \Pi$, by Condition (3), $\sigma(\pi, (V', E'), p) \geq H_\pi(p) - (\sum_{\langle g, \pi \rangle \in E} \lambda_\pi^g + \delta_s)$. Because no allocation is any better for an agent than its optimal allocation, $\sigma(\pi, (V^*, E^*), p) \leq H_\pi(p)$. Thus,

$$\sigma(\pi, (V^*, E^*), p) - \sigma(\pi, (V', E'), p) \quad \leq \quad \sum_{\langle g, \pi \rangle \in E} \lambda_\pi^g + \delta_s. \tag{5}$$

Equations (2)–(5) together imply that $\text{value}((V^*, E^*)) - \text{value}((V', E')) \leq \sum_{\pi \in \Pi} [\sum_{\langle g, \pi \rangle \in E} \lambda_\pi^g + \delta_s] + |C| \delta_b$. Condition (1) implies that each sum term in Equation (2) is nonnegative, hence $\text{value}((V', E')) \geq 0$. As noted, $(V', E')$ is feasible.
$\square$

### A.4 Proof of Theorem 5

In proving the theorem, we refer to the C-level and S-level of producers in a network, as defined in the beginning of Section A.

A task dependency network $(V, E)$ is characterized by the following parameters:

- $\phi$: the maximum C-level of any producer in the network,

- $\Upsilon$: the maximum number of input goods for any producer,

- $R$: the maximum consumer value, $\max_{\langle g, c \rangle \in E | c \in C} v_c(g)$.

**Lemma 23** *In a run of SAMP-SB for network $(V, E)$, no agent places a buy offer above $R + 2\phi\delta_b$.*

*Proof.* Consumers never offer above their valuation, which is bounded by $R$. We prove by induction on the producer C-level that no producer at C-level $k$ places a buy offer above $R + 2k\delta_b$.

Suppose that a producer $\pi$ at C-level one places an offer to buy an input $g$ at price $\beta > R + 2\delta_b$. Since it always increments buy offers by $\delta_b$, this means at some previous time it submitted a buy offer for $g$ at price $R + \delta_b < \beta' \le R + 2\delta_b$. At that time, it must have been losing $g$, else it would not be bidding. But then the ask quote of $g$ must have been greater than $R$, and $\pi$ offered greater than $R$ for its output. Since only a consumer will offer to buy for the output of a producer at C-level one, $\pi$ must lose its output. Because offers are nondecreasing, this situation is permanent, and hence $\pi$ never again raises an input offer, contrary to our supposition. Thus a C-level one producer will never place a buy offer above $R + 2\delta_b$.

For the inductive step, we assume that no producer at any C-level $i$, where $i < k$, places a buy offer above $R + 2i\delta_b$. Thus, no producer at C-level $k$ can win its output offer for more than $R + 2(k-1)\delta_b$. Applying reasoning analogous to the base case (C-level one), we see that no producer at C-level $k$ places a buy offer above $R + 2k\delta_b$. Because $k \le \phi$ for all producers, the lemma follows immediately. □

**Lemma 24** *No agent places more than $\Upsilon(R + 2\phi\delta_b)/\delta_b + \Upsilon$ buy offers.*

*Proof.* Since consumers offer at most $R$ and increase offers by at least $\delta_b$, they place offers at most $R/\delta_b$ times. A producer initially places at most $\Upsilon$ buy offers for its inputs. According to Lemma 23 and the producer bidding policy, a producer subsequently offers no higher than $R + 2\phi\delta_b$ in increments of $\delta_b$ for each of a maximum of $\Upsilon$ inputs. □

**Theorem 5** *SAMP-SB reaches quiescence after a finite number of bids have been placed.*

*Proof.* By Lemma 24, a finite number of buy offers are placed. We need show only that producers place a finite number of sell (output) offers to establish that a finite number of total bids are placed.

A producer will change its output offer only if: 1) the price of an input changes, 2) the ask price of an input changes, or 2) it loses an an offer for a good that it was previously winning. An unchanged input offer can switch between winning and losing at most once without the price changing. Similarly, an unchanged input offer can switch winning state at most once without the ask price changing. Hence, it is sufficient to show that the price and ask price of each of a producer's input goods change at most a finite number of times. We prove by induction on the producer S-level that the price and ask quotes of an input good to a producer at S-level $k$ changes a finite number of times.

Only a producer with no input places an output offer for a input good $g$ to a producer at S-level one, and producers with no input place only one offer each. Hence, the price or ask price of $g$ change only in response to a change in a buy offer for $g$. But by Lemma 24, the number of buy offer changes for $g$ is finite.

Now assume that all producers at all S-levels less than $k$ place a finite number of output offers. For a good $g$ which is an input for a producer $\pi$ at S-level $k$, the number of output offer changes is finite. Again the number of input offers for $g$ must be finite. Since the number of input and output offers for $g$ is finite, $\pi$ places a finite number of output offers. □

## A.5 Proof of Theorem 8

In proving the theorem, we refer to the C-level of producers in a network, as defined in the beginning of Section A. For reference, quasi-quiescence is described in Definition 5.

**Lemma 25** *If a run of SAMP-SB is in a quasi-quiescent state during the time interval $[t, t']$ then no inactive producer changes an offer for an input good in the time interval $[t, t' + \varepsilon]$, where $\varepsilon$ is the smallest period of time an agent requires to update a bid in response to a price quote.*

*Proof.* By definition of quasi-quiescence, during the interval $[t, t']$, no consumer or active producer changes any offer. Thus, a simple induction on the C-level of the inactive producers shows that any producer that is inactive at time $t$ would not win its output during $[t, t']$, hence inactive producers remain inactive during this interval. But a producer that is inactive during $[t, t']$ would not change its input offer during $[t, t' + \varepsilon]$. □

**Lemma 26** *If a run of SAMP-SB is in a quasi-quiescent during the time interval $[t, t']$, then it is quasi-quiescent during the time interval $[t, t' + \varepsilon]$, where $\varepsilon$ is the smallest period of time an agent requires to update a bid in response to a price quote.*

*Proof.* Assume, contrary to that we wish to prove, that a run of SAMP-SB is quasi-quiescent during $[t, t']$ but not during time $[t', t' + \varepsilon]$. Let $a$ be a consumer or active producer that will change an offer in $[t', t' + \varepsilon]$.

If $a$ is a consumer, then $a$ would only change an offer if it lost some offer it was previously winning in quasi-quiescence. If $a$ is a producer, it must be feasible, otherwise it would change its input offer (because it is active) violating quasi-quiescence. Since $a$ is feasible, it would change an offer only if it loses an input it was previously winning, or the price of one of its inputs increases. In any of these cases, $a$ either loses a buy offer it was previously winning, or the price of one of its buy offers increased. For one of these to occur, it must be that at time $t'' \in [t, t']$, some agent either 1) changed its own winning output offer or 2) changed its input offer. But the definition of quasi-quiescence precludes #1, and Lemma 25 and the definition of quasi-quiescence preclude #2. This gives us a contradiction, proving the lemma. □

**Lemma 27** *If a run of SAMP-SB is in a quasi-quiescent state at time $t$, then it is quasi-quiescent at all times $t' > t$.*

*Proof.* By Lemma 26 we can conclude that we have quasi-quiescence in the interval $[t, t+\varepsilon]$, then further extend that interval by $\varepsilon$, and so on indefinitely. □

**Lemma 28** *If a run of SAMP-SB is in a quasi-quiescent at time $t$, then a producer that is inactive at time $t$ is inactive at time $t+\varepsilon$, and a producer that is active at time $t$ is also inactive at time $t+\varepsilon$, where $\varepsilon$ is the smallest period of time an agent requires to update a bid in response to a price quote. Furthermore, $p(g)$ does not change for any good $g$ at time $t+\varepsilon$*

*Proof.* Since an agent cannot lower its offers, the only way for an inactive producer $\pi$ to become active is for some other agent to raise its buy offer. By Lemma 27 and the definition of quasi-quiescence, only inactive producers will change any offers after $t$, and by Lemma 25 no inactive producer will change its input offers. But then $\pi$ remains inactive.

Since offers do not decrease, an active producer $\pi$ can become inactive only by increasing its offer for its output. But $\pi$ will do this only if the prices on its inputs increase. Since we have a quasi-quiescent state, this can happen only if an inactive producer $\pi'$ changes its offer for its output $g$. But since $\pi'$ is inactive, a change to its offer for $g$ can cause only $\alpha(g)$ to change. Since active producers are feasible (otherwise they would want to change their bids, violating quasi-quiescence), $\pi$ is not losing a buy offer for $g$ at time $t$. Therefore, $\pi$ does not respond to changes in $\alpha(g)$, hence does not change its output offer and will remain active. □

**Theorem 8** *If a run of SAMP-SB reaches a quasi-quiescent state, then it remains in a quasi-quiescent state. Furthermore, neither the allocation nor the prices $p$ subsequently change.*

*Proof.* The theorem follows directly from Lemmas 27 and 28. □

### A.6 Proof of Theorem 10

In proving the theorem, we refer to the C-level of producers in a network, as defined in the beginning of Section A.

A given run of SAMP-SB in network $(V, E)$ is characterized by the following parameters:

- $\phi$: the maximum C-level of any producer in the network,

- $\Upsilon$: the maximum number of input goods for any producer,

- $R$: the maximum consumer value, $\max_{\langle g,c \rangle \in E|c \in C} v_c(g)$.

**Theorem 10** *SAMP-SB reaches a quasi-quiescent state after a number of bids bounded by a polynomial of the size of the network and the value of the maximum consumer value have been placed by consumers and active producers.*

*Proof.* SAMP-SB is guaranteed to reach a quasi-quiescent state (Theorem 5 and Observation 7). By Lemma 24, the number of buy offers is bounded by a polynomial in the value of $R$, hence we need only be concerned with the number of sell offers placed. Since the prices of buy offers increase by at least $\delta_b$, a producer's perceived cost for any good must rise by at least $\delta_b$, so will increase its sell offer by no less than $\delta_b$. Also, a producer will increase its sell offer by no less than $\delta_s$, as required by the auction. Hence, Lemma 23 implies that an active producer will become permanently inactive after it places at most $(R + 2\phi\delta_b)/[\max(\delta_b, \delta_s)]$ output offers. □

## A.7 Proof of Theorem 11

In proving the theorem, we refer to the conditions for a $\lambda$-$\delta$-competitive equilibrium (Definition 4).

**Lemma 29** *When SAMP-SB reaches quiescence in network $(V, E)$ then each consumer obeys the $\lambda$-$\delta$-competitive equilibrium conditions (Conditions (1) and (2)).*

*Proof.* Since each consumer maintains at most a single winning offer for a good that gives it nonnegative surplus, it obeys Condition (1).

Let the final prices and allocation be $p$ and $(V', E')$, respectively. Assume, contrary to Condition (2), that $\sigma(c, (V', E'), p) < H_c(p) - \delta_b$ for some consumer $c$. Let $g^*$ be a surplus-maximizing good for $c$ at $p$.

If $c$ does not buy a good, then $p(g^*) + \delta_b < v_c(g^*)$ (otherwise it would have placed and won an offer for $g^*$) and $\sigma(c, (V', E'), p) = 0$. Noting also that $H_c(p) = v_c(g^*) - p(g^*)$, algebraic manipulation gives us $\sigma(c, (V', E'), p) > H_c(p) - \delta_b$, which is a contradiction.

Thus, $c$ buys one good $g'$ such that

$$v_c(g') - p(g') \quad < \quad v_c(g^*) - p(g^*) - \delta_b. \tag{6}$$

Let $\hat{p}(g^*)$ and $\hat{p}(g')$ be the prices for $g^*$ and $g'$ when $c$ placed its final offer for $g'$. Since $c$ offers $\hat{p}(g') + \delta_b$ for $g'$, and since $c$ won this offer at $p(g')$, we have

$$\hat{p}(g') + \delta_b \geq p(g'). \tag{7}$$

Since prices do not decrease, we have

$$\hat{p}(g^*) \leq p(g^*). \tag{8}$$

Substituting Equations (7) and (8) into the left and right sides, respectively, of Equation (6) gives us

$$v_c(g') - (\hat{p}(g') + \delta_b) \quad < \quad v_c(g^*) - (\hat{p}(g^*) + \delta_b).$$

But the consumer bidding policy specifies that $c$ would have bid for $g^*$, rather than $g'$ at prices $\hat{p}$, which is a contradiction. Thus each consumer obeys Condition (2). □

**Lemma 30** *If SAMP-SB reaches quiescence in network $(V, E)$ such that no inactive producer buys a positive-price input, then each producer obeys the $\lambda$-$\delta$-competitive equilibrium conditions (Conditions (1) and (3)), with $\lambda_\pi^g = \max(\alpha(g) - p(g), \delta_b)$.*

*Proof.* The bidding policy ensures that each producer $\pi$ sells its output $g_\pi$ only at a nonnegative surplus, and the lemma conditions directly imply that $\pi$ has zero surplus if it does not sell $\pi$. Thus $\pi$ obeys Condition (1).

The producer bidding policy guarantees that $\pi$ is feasible in quiescence.

Let the final prices be $p$ and allocation be $(V', E')$. If $H_\pi(p) > \sum_{\langle g,\pi \rangle \in E} \lambda_\pi^g + \delta_s$ in quiescence, then $H_\pi(p) = p(g_\pi) - \sum_{\langle g,\pi \rangle \in E} p(g) > \sum_{\langle g,\pi \rangle \in E} \lambda_\pi^g + \delta_s$. Thus $p(g_\pi) > \sum_{\langle g,\pi \rangle \in E} p(g) + \sum_{\langle g,\pi \rangle \in E} \lambda_\pi^g + \delta_s$. The producer bidding policy ensures that $\pi$ offers no more than $\sum_{\langle g,\pi \rangle \in E} p(g) + \sum_{\langle g,\pi \rangle \in E} \lambda_\pi^g + \delta_s$ for $g_\pi$, so it must be winning $g_\pi$ at a profit. Thus $\sigma(\pi, (V', E'), p) = H_\pi(p)$ and Condition (3) holds.

If instead $H_\pi(p) \le \sum_{\langle g,\pi \rangle \in E} \lambda_\pi^g + \delta_s$, then since $\sigma(\pi, (V', E'), p) \ge 0$, Condition (3) holds. □

**Theorem 11** *The prices and allocation determined in quiescence by the SAMP-SB protocol is a $\lambda$-$\delta$-competitive equilibrium, with $\lambda_\pi^g = \max(\alpha(g) - p(g), \delta_b)$, iff no inactive producer buys any positive-price input.*

*Proof. Case only if:* Condition (1) of $\lambda$-$\delta$-competitive equilibrium (Definition 4) fails if an inactive producer buys any positive-price input.

*Case if:* Lemmas 29 and 30 show that the consumers and producers, respectively, obey the $\lambda$-$\delta$-competitive equilibrium conditions (Conditions (1)–(3)). The (M+1)st-price auction rules ensure Condition (4). All conditions of $\lambda$-$\delta$-competitive equilibrium are met. □

## A.8 Proof of Theorem 12

In proving the theorem, we refer to the conditions for a $\lambda$-$\delta$-competitive equilibrium (Definition 4).

**Lemma 31** *If $\alpha(g) - p(g) > \delta_b$ for any good $g$ in a quiescent state of SAMP-SB for network $(V, E)$, then no agent wins an offer for $g$.*

*Proof.* Assume, contrary to which we wish to prove, that $\alpha(g) - p(g) > \delta_b$ and some agent is winning an offer for $g$, in quiescence of SAMP-SB. Either a buy offer or a sell offer sets $\alpha(g)$.

*Case 1: An agent sets $\alpha(g)$ with a buy offer.* According to the SAMP-SB bidding policies, an agent will increase a buy offer only if it is losing that offer. An agent will win any offer for $g$ at a price above $p(g)$. A producer increases its buy offer in increments of $\delta_b$ and a consumer offers at most $p(g) + \delta_b$. In either case, an agent will place a buy offer no higher than $p(g) + \delta_b$ for $g$. But then $\alpha(g) \le p(g) + \delta_b$, which is a contradiction.

*Case 2: An agent sets $\alpha(g)$ with a sell offer.* As with Case 1, there are no buy offers higher than $p(g) + \delta_b$, hence every buy offer is strictly below $\alpha(g)$. Recall that, if there are $M$ sell offers, the $M$th highest offer determines $\alpha(g)$. Then since there are no buy offers at or above $\alpha(g)$, it must be that all sell offers are at or above $\alpha(g)$. But then all sell offers are strictly above all buy offers and no agent wins an offer for $g$, which is a contradiction.

Since each case gives us a contradiction, it must be the case that no agent wins an offer for $g$ when $\alpha(g) - p(g) > \delta_b$. □

**Lemma 32** *If $(V', E')$ is in $\lambda$-$\delta$-competitive equilibrium at prices $p$, in quiescence of SAMP-SB for network $(V, E)$, then there exist prices $p'$ such that $(V', E')$ is also in $\lambda$-$\delta$-competitive equilibrium at $p'$, with $\lambda_\pi^g = \delta_b$ for all producers $\pi$ and goods $g$.*

*Proof.* We specify $p'$ as follows: if $\alpha(g) > p(g) + \delta_b$, then $p'(g) = \alpha(g)$, otherwise $p'(g) = p(g)$. We will show that all the conditions of $\lambda$-$\delta$-competitive equilibrium hold with $\lambda_\pi^g = \delta_b$. Because we are considering the same allocation, the goods are still in material balance so Condition 4 still holds at $p'$.

Consider an agent $a$ such that $p'(g) = p(g)$ for all adjacent goods $g$. Clearly, $H_a(p') = H_a(p)$ and $\sigma(a, (V', E'), p') = \sigma(a, (V', E'), p)$. We then have $\sigma(a, (V', E'), p') \geq 0$ (Condition 1), and the surplus bound is met for consumers (Condition 2) since these hold at $p$. If $a$ is a producer, then for any input $g$, $\alpha(g) - p(g) \leq \delta_b$ since $p'(g) = p(g)$. Hence, we have the following bound on its perceived cost for $g$: $\hat{p}_a(g) \leq p(g) + \delta_b$. As a result, the producer bidding policies imply that $\sigma(a, (V', E'), p) \geq H_\pi(p) - (\sum_{\langle g, \pi \rangle \in E} \delta_b + \delta_s)$. Therefore $\sigma(a, (V', E'), p') \geq H_\pi(p') - (\sum_{\langle g, a \rangle \in E} \lambda_a^g + \delta_s)$ and and the producer surplus bound (Condition 3) holds with $\lambda_a^{g'}$ for all inputs $g'$.

Now consider an agent $a$ adjacent to a good $g$ with $p'(g) = \alpha(g)$. By Lemma 31, $a$ does not win an offer for $g$, so $\sigma(a, (V', E'), p') = \sigma(a, (V', E'), p)$, implying $\sigma(a, (V', E'), p') \geq 0$ (Condition 1). If $a$ is a consumer, since $p'(g) \geq p(g)$, and since $a$ does not win $g$, we have $H_a(p') = H_a(p)$, so the surplus bound is met for consumers (Condition 2).

If $a$ is a producer, then since it did not win $g$, it must not have won any good (according to the $\lambda$-$\delta$-competitive equilibrium conditions and Theorem 11), implying $\sigma(a, (V', E'), p') = 0$. The producer bidding policy specifies that $a$ offered a price at most $\beta = \kappa_a + \sum_{\langle g', a \rangle \in E} \max(\alpha(g'), \ p(g') + \delta_b) + \delta_s$ for its output $g_a$. Since $a$ did not win $g_a$, it must be that $\alpha(g_a) \leq \beta$. But by the way $p'$ is constructed, $p'(g_a) \leq \alpha(g_a)$ and $p'(g') + \delta_b \geq \max(\alpha(g'), \ p(g') + \delta_b)$, giving us $p'(g_a) \leq \alpha(g_a) \leq \beta \leq \kappa_a + \sum_{\langle g', a \rangle \in E}(p'(g') + \delta_b) + \delta_s$. If $a$ would optimize at $p'$ by being active, we have $H_a(p') = p'(g_a) - \kappa_a - \sum_{\langle g', a \rangle \in E} p'(g') \leq \sum_{\langle g', a \rangle \in E} \delta_b + \delta_s$. But since $\sigma(a, (V', E'), p') = 0$ it follows that $\sigma(a, (V', E'), p') \geq H_a(p') - (\sum_{\langle g', a \rangle \in E} \delta_b + \delta_s)$. If, on the other hand, $a$ would optimize at $p'$ by being inactive at, $\sigma(a, (V', E'), p') = H_a(p')$. In either case, the surplus bound is met for producers (Condition 3) with $\lambda_a^{g'} = \delta_b$ for all inputs $g'$. $\square$

**Theorem 12** *If $(V', E')$ is a $\lambda$-$\delta$-competitive equilibrium computed by SAMP-SB then $(V', E')$ has a nonnegative value that differs from the value of an efficient allocation by at most $\sum_{\pi \in \Pi}(|\{\langle g, \pi \rangle \in E\}| \ \delta_b + \delta_s) + |C|\delta_b$.*

*Proof.* By Lemma 32, there is a $\lambda$-$\delta$-competitive equilibrium for $(V', E')$ with $\lambda_\pi^g = \delta_b$ for all producers $\pi$ and goods $g$. With $\delta_b$ substituted for $\lambda_\pi^g$ in the equation from Theorem 3, we have proved the present theorem. $\square$

## A.9 Proof of Theorem 13

In proving the theorem, we refer to the S-level of producers in a network, as defined in the beginning of Section A.

**Theorem 13** *The quiescent state of SAMP-SB is a $\lambda$-$\delta$-competitive equilibrium for a tree.*

*Proof.* We prove, by induction on the S-level of producers, that no producer changes its initial output offer. Since buy offers never decrease, it follows that, if a producer is winning its output, it will not lose its output at any successive state of the run of the protocol. Since a producer bids for its inputs only when winning its output, no inactive producer will buy any positive-price output and the present theorem follows from Theorem 11.

*Basis case:* The bidding policy specifies that a producer at S-level zero never changes its initial output offer.

*Inductive case:* Assume that no producer at S-level less than $k$ changes its initial output offer to show that a producer $\pi$ at S-level $k$ never changes its initial output offer. Consider input good $g$ with $M$ sell offers and the lowest sell offer $\beta$. Since the network is a tree, $\pi$ is the only agent that places buy offers for $g$. Producer $\pi$ initially offers zero for $g$, and so long as it offers less than $\beta$ it loses its offer, and $\hat{p}_\pi(g) = \alpha(g)$. While this holds, $\alpha(g)$, defined as the $M$th highest price, is the lowest sell offer, hence $\hat{p}_\pi(g) = \beta$. As soon as $\pi$ offers $\beta$ or greater for $g$ it will win its offer, and then $\hat{p}_\pi(g) = p(g)$. When this holds, $p(g)$, defined as the $M+1$st highest price, is the lowest sell offer, hence $\hat{p}_\pi(g) = \beta$. We conclude that $\hat{p}_\pi(g)$ never changes for any input $g$, hence $\pi$ never changes its initial output offer.

We have proven that no producer changes its initial output offer, and by the argument above, the theorem is proven. $\square$

## A.10 Proof of Theorem 14

**Theorem 14** *The quiescent state of safe SAMP-SB is a $\lambda$-$\delta$-competitive equilibrium for a network with no input complementarities.*

*Proof.* We will show that that no inactive producer buys its input at a positive price in quiescence of safe SAMP-SB. Since the properties of safe SAMP-SB in quiescence are the same as in SAMP-SB, the present theorem then follows from Theorem 11. Assume, to the contrary, that, in quiescence, producer $\pi$ wins its input $g$ at a positive price but loses its offer for output $g_\pi$.

Let $\beta$ be the price of the final offer by $\pi$ for $g$, $p(g) > 0$ be the final price of $g$, and $\hat{p}_\pi(g)$ be the final perceived cost of $g$ to $\pi$. Since $\pi$ wins $g$ in quiescence, $\hat{p}_\pi(g) = p(g)$. Let $\beta'$ be the price of the second to last offer from $\pi$. Immediately before $\pi$ places offer $\beta$, let $\hat{p}'_\pi(g)$ be the perceived price of $g$ to $\pi$ and $p'(g)$ be the price component from the price quote for $g$. According to the bidding policy, $\beta = \beta' + \delta_b$. Since $\pi$ offers $\beta$ only if it loses $g$ with offer $\beta'$, it must be that $\beta' \leq p'(g)$, hence $\beta \leq p'(g) + \delta_b$. Furthermore, since $\pi$ loses with offer $\beta'$, we have $\hat{p}'_\pi(g) \geq p'(g) + \delta_b$. Because we assume that $\pi$ wins $g$ in quiescence, it must be that $p(g) \leq \beta$, hence $p(g) \leq p'(g) + \delta_b$. It follows that, since $\hat{p}_\pi(g) = p(g)$ and $\hat{p}'_\pi(g) \geq p'(g) + \delta_b$, we have $\hat{p}_\pi(g) \leq \hat{p}'_\pi(g)$.

According to the safe SAMP-SB bidding policies, $\pi$ offers $\beta$ for $g$ only if it is first winning $g_\pi$ with offer price $\hat{p}'_\pi(g)$. Since $\hat{p}_\pi(g) \leq \hat{p}'_\pi(g)$, $\pi$ its offer for $g_\pi$ is the same in quiescence as when it had placed $\beta$ for $g$. But since no offers from any agent decrease, $\pi$ must continue to win its final offer for $g_\pi$, contradicting the assumption that $\pi$ loses $g_\pi$ in quiescence. Thus, $\pi$ does not win its input at a positive price if it is inactive, and the quiescent state of safe SAMP-SB is a $\lambda$-$\delta$-competitive equilibrium. $\square$

## A.11 Proof of Theorem 15

In proving the theorem, we refer to the C-level and S-level of producers in a network, as defined in the beginning of Section A.

**Theorem 15** *If $(V, E)$ is a polytree with a solution that assigns good $g$ to consumer $c$, then given all other costs and values, there exists a value $v_c(g)$ such that SAMP-SB is guaranteed to converge to a valid solution $(V', E')$ for $c$.*

*Proof.* For convenience, denote $\max\left(\sum_{\pi \in \Pi \cap V'} \kappa_\pi, \max_{c' \in C, \langle g', c'\rangle \neq \langle g, c\rangle} v_{c'}(g')\right)$ as $\gamma$. We show that the theorem holds for:

$$v_c(g) = [\gamma + (2\delta_b + \delta_s)|\Pi|]\,|\Pi| + \delta_b.$$

We need to show that SAMP-SB cannot reach a state in which $p(g) > v_c(g) - \delta_b$ and $c$ is not winning $g$, because then $c$ would stop bidding for $g$ and the desired solution would not form.

First, observe that for any consumer $c'$ and any good $g'$ such that $\langle g', a\rangle \neq \langle g, c\rangle$, $c'$ will not offer above $\gamma$ for $g'$, by construction.

Now, consider a producer $\pi$ such that there is no directed path from $\pi$ to $g$ through the output of $\pi$. We show, by induction on the C-level of producers, that no such producer offers higher than $\gamma + \delta_b d_\pi$, where $d_\pi$ is the C-level of $\pi$, for one of its inputs. For the basis case, such a producer $\pi$ at C-level one cannot win an output offer above $\gamma$ (by the definition of $\gamma$). $\pi$ increases its input offers in increments of $\delta_b$, so to offer $\beta' > \gamma + \delta_b$, on any input $g'$, it must first offer $\beta$, where $\gamma < \beta \leq \gamma + \delta_b$ for that input. $\pi$ will only offer $\beta'$ if it is losing $\beta$ but winning its output offer. But if $\pi$ is losing $\beta$, we must have $p(g') \geq \beta$, so $\pi$ must be offering more than $\gamma$ for its output. But then it cannot be winning its output, hence would not offer $\beta'$ for $g$. Thus $\pi$ at C-level one does not offer more than $\gamma + \delta_b$ for any input, establishing the base case. Now, assume that the property holds for every such producer at C-level less than $k$ to show that it holds for producer $\pi$ at C-level $k$. Given the inductive assumption, it must be that $\pi$ cannot win its output for more than $\gamma + \delta_b(k-1)$. By an argument similar to the basis case, $\pi$ does not offer above $\gamma + \delta_b k$ for its input, proving the inductive case. Since $d_\pi \leq |\Pi|$, then no such producer offers higher than $\gamma + \delta_b|\Pi|$ for its input.

For any producer $\pi \in \Pi \cap V'$, denote as $I_\pi$ the maximum number of producers, other than $\pi$, in the subgraph of $(V, E)$, rooted at $\pi$. Now we show by induction on the S-level, that a producer $\pi$ on a directed path to $g$ offers no more than $[\gamma + \delta_b(|\Pi| + d_\pi) + \delta_s(d_\pi - 1)]I_\pi + \delta_s$ for its output, where $d_\pi$ is the S-level of $\pi$. For the basis case, consider such a producer $\pi$ at S-level one, offering to buy some $g'$. No consumer offers above $\gamma$ for $g'$. Because $(V, E)$ is a polytree, any other producer $\pi'$ that offers to buy $g'$ is not on a directed path to $g$, hence offers at most $\gamma + \delta_b|\Pi|$ to buy $g'$. Any producer that offers to sell $g'$ must have no inputs, hence offers no more than $\gamma$ for $g'$. Hence $\pi'$ can successfully buy $g'$ with a offer no higher than $\gamma + \delta_b(|\Pi| + 1)$, thus will offer no higher than this amount for $g'$. Since the number of inputs to $\pi$ is equal to $I_\pi$, it will offer no more than $(\gamma + \delta_b(|\Pi| + 1))I_\pi + \delta_s$ for its output, and the basis case is proven. Now, assume that the property holds for every such producer at S-level less than $k$ to prove that it holds for producer $\pi$ offering to buy some $g'$ at S-level $k$. By the inductive assumption, no producer $\bar\pi$ offers to sell $g'$ for more than $(\gamma + \delta_b(|\Pi| + k - 1) + \delta_s(k - 2))I_{\bar\pi} + \delta_s$. As in the basis case, no consumer offers more than $\gamma$ for $g'$ and any producer other than $\pi$ will offer no more than $\gamma + \delta_b|\Pi|$ to buy $g'$. Hence, $\pi$ will offer at most $(\gamma + \delta_b(|\Pi| + k - 1) + \delta_s(k - 2))I_{\bar\pi} + \delta_s + \delta_b$ to buy $g'$, and for its output will offer at most

$$\left[\sum_{\langle \bar\pi, g'\rangle p \in E \,|\, \langle g', \pi\rangle \in E} (\gamma + \delta_b(|\Pi| + k - 1) + \delta_s(k - 2))I_{\bar\pi} + \delta_s + \delta_b\right] + \delta_s \leq$$

$$[\gamma + \delta_b(|\Pi| + k) + \delta_s(k - 1)]I_\pi + \delta_s,$$

proving the inductive case. Since $I_\pi \leq |\Pi|$ and $d_\pi \leq |\Pi|$, then no such producer offers higher than $[\gamma + 2\delta_b|\Pi| + \delta_s(|\Pi| - 1)]|\Pi| + \delta_s \leq [\gamma + (2\delta_b + \delta_s)|\Pi|]\,|\Pi| = v_c(g) - \delta_b$.

We have shown that no agent $a \neq c$ places a buy offer as high as $v_c(g) - \delta_b$ for $g$ and no producer on a directed path to $g$ places a sell offer as high as $v_c(g) - \delta_b$ for $g$. Hence $c$ is the only agent that could possibly offer as high as $v_c(g) - \delta_b$ for $g$. But $c$ will offer this high, if necessary, to win $g$, and will win $g$ if it offers $v_c(g) - \delta_b$ or higher. It follows that $c$ will win $g$ at a price below $v_c(g)$ in quiescence. By Observation 7 and Theorem 16, the state must be a valid solution. □

## A.12 Proof of Theorem 16

**Theorem 16** *If SAMP-SB reaches quasi-quiescence with $p(g) < v_c(g)$ for some $\langle g,c \rangle \in E$, $c \in C$, then the system's state represents a valid solution.*

*Proof.* Because the definition of quasi-quiescence requires that active producers do not change their bids, they must be feasible. All other agents are feasible by definition. The price of an active producer's output good must be no less than the total price of its input goods, otherwise it would increase its output offer, violating quasi-quiescence.

Because $p(g) < v_c(g)$, consumer $c$ must have won its offer for $g$. A consumer bids in such a way that it wins only one unit of one good, and consumers do not change their bids in quasi-quiescence.

Finally, the auction guarantees that there is a one-to-one mapping between successful buy offers and successful sell offers for any good, ensuring material balance.

Thus, each of the constraints for a valid solution is satisfied. □

## A.13 Proof of Theorem 17

**Theorem 17** *If a run of SAMP-SB in $(V, E)$ is in a valid solution state such that:*

- *each consumer $c$ is either winning an offer or $p(g) + \delta_b > v_c(g)$ for all $\langle g,c \rangle \in E$,*

- *all agents have correct beliefs about which goods they are currently winning,*

- *all bids from consumers and active producers have been received in response to the current price quotes,*

- *and no sell offers are lost due to tie breaking,*

*then after the subsequent price quote from each auction, the system will be in a quasi-quiescent state with a valid solution.*

*Proof.* Let the current prices be $p$. The consumer bidding policy dictates that the consumers do not change their offers under the specified conditions. Because we have a valid solution, each producer is feasible and thus will not raise any of its buy offers for inputs. Therefore, no agent changes any buy offers.

An active producer $\pi$ is feasible in a valid solution. Since $\pi$ is winning all of its inputs, it only raises its offer for output $g_\pi$ if $p(g)$ changes for once of its inputs $g$, and will place an offer for its output a price no higher than the sum of its input good prices. By the definition of a valid solution, if $\pi$ is active, then the current price of its output good is no less than sum of the current prices of its inputs. But since $\pi$ won its offer for $g_\pi$, it must have offered no higher than $p(g_\pi)$ for $g_\pi$. Because the previous offer price by $\pi$ for $g_\pi$ is no higher than $p(g_\pi)$, and because the sum of the current prices of its inputs are no higher than $p(g)$, $\pi$ will offer no higher than $p(g_\pi)$ for $g_\pi$.

We have established that no agent changes any buy offers, and no currently active producer places a sell offer above $p(g)$ for any good $g$. We show this implies that, at the next price quotes with prices $p'$, we have $p'(g) = p(g)$. Assume the contrary. Since offers do not decrease, $p'(g) > p(g)$. Since no buy offer and no winning sell offer changed, the price increase is due an updated losing sell offer at price $\beta$, such that $\beta = p'(g)$. But if the agent was losing with a previous offer price of $\beta'$, it must be that $\beta'$ was at least as high as the $(M+1)$st highest offer. Thus $\beta$, being higher, must be strictly higher than the $(M+1)$st highest offer, hence cannot raise the price of $g$. Hence $p'(g) = p(g)$.

Since prices do not change, the temporal-precedence tie breaking ensures that the set of winning buy offers does not change. Additionally, since no winning seller offers above $p(g)$ and no sell offers are currently lost to tie breaking, the set of winning sell offers does not change. Since prices and allocations do not change, no consumer or active producer will change its bids. Furthermore, because the system is in a valid solution state based on the current price quotes, it must be in a valid solution state based on the next price quotes. □

We note that temporal-precedence tie-breaking itself (without the requirement that no tied sell offers are lost) is not sufficient to ensure that the allocation to sellers does not change. If some tied sell offers are lost, it is possible that an active producer could increase its next sell offer price up to the price of its output good. If this occurs, then that producer would lose the tie breaking of its output at the next quote, and the system would not be in quasi-quiescence.